\documentclass[letterpaper]{article} 
\usepackage[]{univvt}  
\usepackage[hyphens]{url}  
\usepackage{graphicx} 
\usepackage{natbib}  
\usepackage{caption} 
\usepackage{algorithm}
\usepackage{algorithmic}
\usepackage{subfig}   
\usepackage{multirow}  
\usepackage{makecell}  
\usepackage{bbding}
\usepackage{xcolor}   
\usepackage{colortbl} 
\usepackage{circledsteps} 
\usepackage{hhline} 
\usepackage{float}
\usepackage{amsmath}
\usepackage{amssymb}
\usepackage{newfloat}
\usepackage{listings}
\DeclareCaptionStyle{ruled}{labelfont=normalfont,labelsep=colon,strut=off} 
\floatstyle{ruled}
\newfloat{listing}{tb}{lst}{}
\floatname{listing}{Listing}

\usepackage{booktabs}

\title{UniVVT: A Unified End-to-End Framework for High-Fidelity Video Virtual Try-on}
\author{
    Yushe Cao\equalcontrib\textsuperscript{\rm 1}
    Shikun Feng\equalcontrib\corresponding\textsuperscript{\rm 2},
    Fei Shen\textsuperscript{\rm 3},
    Haikuo Peng\textsuperscript{\rm 4},
    Jianqiang Xia\textsuperscript{\rm 5},\\
    Yiheng Zhu\textsuperscript{\rm 2},
    Dianxi Shi\corresponding\textsuperscript{\rm 1},
    Chun Yu\corresponding\textsuperscript{\rm 1},
}
\affiliations{
    \textsuperscript{\rm 1}Tsinghua University
    \textsuperscript{\rm 2}Zhongguancun Academy
    \textsuperscript{\rm 3}National University of Singapore\\
    \textsuperscript{\rm 4}National University of Defense Technology
    \textsuperscript{\rm 5}Shanghai Jiao Tong University

    cao-ys23@mails.tsinghua.edu.cn,kunayumi@163.com
}

\begin{document}

\maketitle

\begin{abstract}
    Video Virtual Try-On (VVT) synthesizes a video of a person wearing a target garment while preserving identity, motion, and scene dynamics. Dominant approaches cast VVT as mask-conditioned video inpainting and rely on separate modules for human parsing, pose estimation, and garment warping. This multi-stage design complicates deployment and, more critically, allows errors in explicit geometric priors to propagate irreversibly into the generated video. We present UniVVT, a unified end-to-end framework that reframes VVT as semantically conditioned video generation, eliminating mask, pose, and warping modules at inference. At its core, a scene-task perceiver built on a Multimodal Large Language Model jointly encodes the source video, target garment, and task instruction into compact, task-aware latent tokens, implicitly capturing what to transfer and where and how to transfer it. A lightweight semantic bridge then aligns these tokens with the conditioning space of a diffusion-based video generator, enabling coherent garment transfer. To robustly couple the heterogeneous components, we devise a three-stage progressive training strategy comprising semantic alignment, joint task adaptation, and flexible-resolution refinement. Extensive experiments demonstrate that UniVVT achieves state-of-the-art  performance across multiple benchmarks, validating implicit semantic guidance as a simple and effective alternative to fragile geometric preprocessing for end-to-end virtual try-on.
\end{abstract}


\section{Introduction}
\label{intro}

Video Virtual Try-On (VVT) synthesizes videos of a person wearing a target garment, enabling dynamic garment visualization for online retail, fashion content creation, and digital humans~\cite{jiang2022clothformer,zou2025video,zuo2025dreamvvt}. Compared with image-based virtual try-on~\cite{yang2020towards,zhu2023tryondiffusion,morelli2023ladi,jiang2024fitdit}, VVT must preserve fine-grained garment appearance in every frame while modeling temporally coherent person--garment interactions under motion, deformation, and occlusion. These coupled requirements for spatial fidelity and temporal consistency make VVT substantially more challenging than static try-on.

Most VVT methods inherit the image-based paradigm~\cite{xu2025ootdiffusion,chong2024catvton,wang2025fw,zhang2026unifit}: they cast try-on as temporal inpainting and use auxiliary models to extract clothing-agnostic masks, parsing maps, pose sequences, or DensePose representations before garment warping and synthesis~\cite{fang2024vivid,chong2025catv2ton,li2025magictryon}. This multi-stage pipeline has three coupled limitations. First, errors from explicit geometric cues propagate to outputs and cause source residuals, incomplete garment replacement and boundary artifacts under occlusions or large motions (Figure 2). Second, many dedicated sub-models increase computational overhead and deployment complexity. Third, disjoint stage-wise optimization stops the generator from compensating upstream errors.
\begin{figure}[t]
  \centering
  \includegraphics[width=\linewidth]{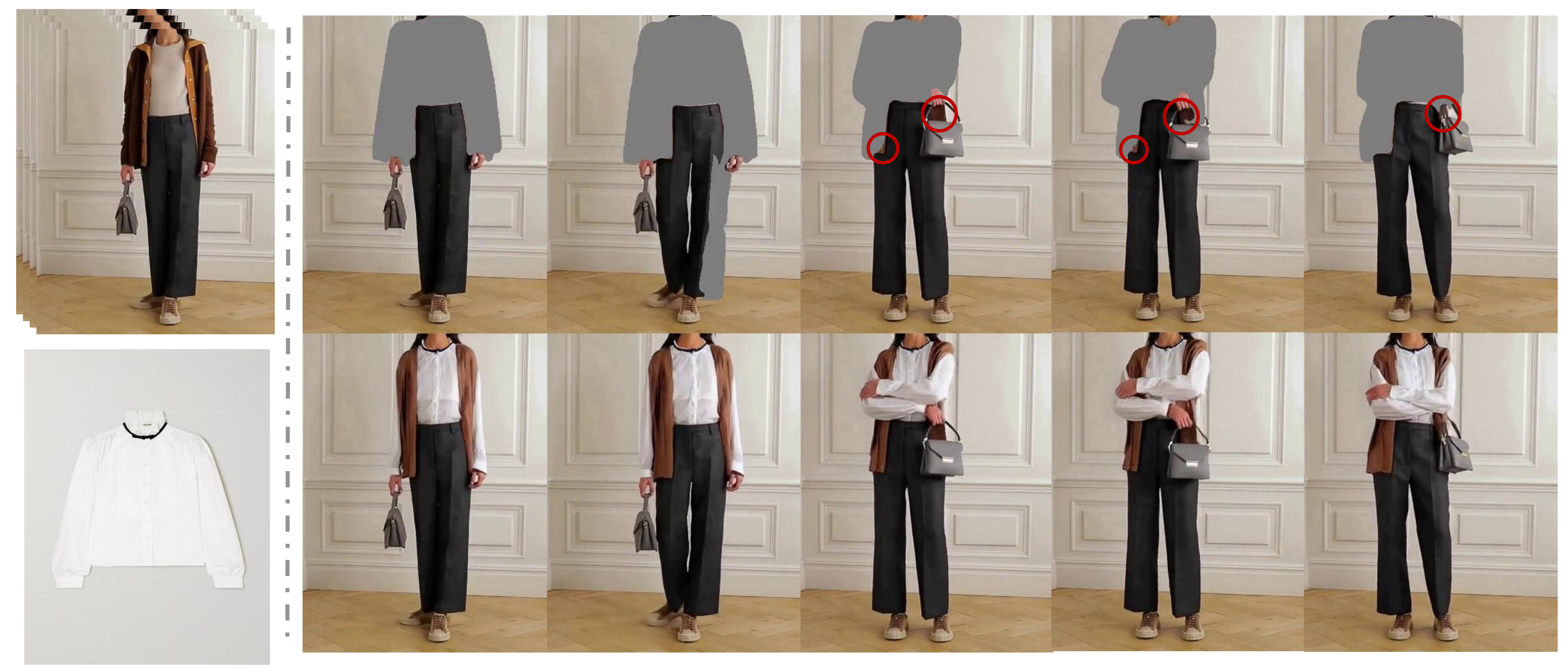}
  \caption{Failure cases caused by inaccurate masks. Leaked details from the original outerwear (top) contaminate the generated blouse (bottom), exposing the fragility of preprocessing-dependent VVT pipelines.}
    \label{fig:issue}
\end{figure}

We propose UniVVT, a unified end-to-end framework that instead formulates VVT as semantically conditioned video generation. Our key insight is that \emph{what} to transfer and \emph{where} and \emph{how} to transfer it can be encoded implicitly by multimodal semantics rather than prescribed through masks, poses, or warping fields. UniVVT realizes this idea through implicit guidance. A Multimodal Large Language Model (MLLM)~\cite{Qwen3-VL} serves as a Scene-Task Perceiver, jointly encoding the source video, target garment, and task instruction into compact task-aware latent tokens. A lightweight Semantic Bridge projects these tokens into the conditioning space of a diffusion-based video generator. UniVVT thereby integrates task perception, semantic alignment, and synthesis without external parsing, pose, or warping modules at inference. Additionally, we introduce a three-stage progressive training strategy to bridge representation gaps among heterogeneous components and stably train a high-fidelity try-on generator. Evaluations on benchmarks including ViViD-S~\cite{fang2024vivid} verify UniVVT’s competitive performance on both image and video virtual try-on. Removing fragile geometric preprocessing further simplifies deployment.

Our contributions are summarized as follows:
\begin{itemize}
    \item We reformulate VVT as semantically conditioned video generation and propose UniVVT, a unified end-to-end framework that requires no external masks, pose estimation, or garment warping at inference.
    \item We introduce an implicit guidance mechanism comprising an MLLM-based Scene-Task Perceiver and a lightweight Semantic Bridge, which translate raw multimodal inputs into task-aware conditions for coherent try-on generation.
    \item We develop a three-stage progressive training strategy to stabilize adaptation across heterogeneous pretrained components. Comprehensive experiments on commonly used benchmarks validate the effectiveness and generality of the proposed framework.
\end{itemize}

\section{Related Work}

\subsection{From Image to Video Virtual Try-On}
Image-based virtual try-on has evolved from explicit warp-and-render pipelines~\cite{yang2020towards,minar2020cp,ge2021parser} to diffusion-driven conditional generation~\cite{yang2024d,velioglu2024tryoffdiff,xing2025tryon,zhang2026unifit,cao2026multivariate}. Early methods align garments with target bodies via geometric transformations or learned flow fields before feeding them into generative networks for final synthesis~\cite{bookstein2002principal,han2019clothflow,ge2021parser}. Diffusion models greatly boost visual realism and garment fidelity~\cite{zhu2023tryondiffusion,fang2024pg,song2025image}. Representative approaches including OOTDiffusion~\cite{xu2025ootdiffusion}, IDM-VTON~\cite{choi2024improving}, StableVITON~\cite{kim2024stableviton}, and FitDit~\cite{jiang2024fitdit} further strengthen garment conditioning via tailored attention modules and feature injection. Nevertheless, most existing methods rely heavily on explicit spatial priors such as parsing masks, human poses, and warped garments, making synthesis vulnerable to preprocessing errors.

Video virtual try-on inherits and exacerbates these limitations along the temporal dimension. Early pipelines perform frame-wise warping and rendering~\cite{zhong2021mv,dong2019fw}, while ClothFormer~\cite{jiang2022clothformer} captures cross-frame dependencies after garment deformation. Recent diffusion-based methods enhance temporal consistency using spatio-temporal attention~\cite{fang2024vivid,zou2025video,zheng2024dynamic}, feature concatenation~\cite{chong2025catv2ton,li2025magictryon}, staged keyframe synthesis, motion-aware modeling\cite{cao2026dual}, and positional encoding~\cite{pan2025once,zuo2025dreamvvt}. Despite progress, prevailing solutions remain formulated as preprocessing-dependent video inpainting~\cite{blattmann2023align,karras2024fashion}: temporally consistent masks and poses predefine where and how edits are applied prior to generation. This causes upstream errors to propagate over frames, and disjointly optimized modules increase overall system complexity. UniVVT breaks this paradigm by replacing explicit geometric intermediates with task-aware semantic conditioning.
\subsection{LLMs and MLLMs for Generative Guidance}
Large Language Models (LLMs) and Multimodal Large Language Models (MLLMs) provide strong instruction understanding and multimodal reasoning capabilities~\cite{qwen3.5,Qwen3-VL,singh2025openai,deng2025exploring,comanici2025gemini,cheng2026conditional}. Existing image-editing systems use language models to construct editing instructions or translate user intent into conditions for diffusion models~\cite{brooks2023instructpix2pix,taseerxmgie}. In video generation, LLMs and MLLMs have primarily served as instruction refiners, storyboard planners, or high-level coordinators~\cite{wangamagic,yuan2025instruction,zhao2025omnialign,mou2025instructx}. Their latent representations, however, remain underexplored as direct conditioning signals for fine-grained, temporally structured editing. UniVVT employs the MLLM as a scene-task perceiver rather than a text generator: it jointly encodes the source video, target garment, and task instruction into compact latent tokens, which a lightweight semantic bridge maps to the conditioning space of a video diffusion model. This design turns multimodal understanding into implicit generative guidance and removes the need to specify the edit through masks, poses, or warping fields at inference.
\begin{figure*}[t]
\centering
  \includegraphics[width=\textwidth]{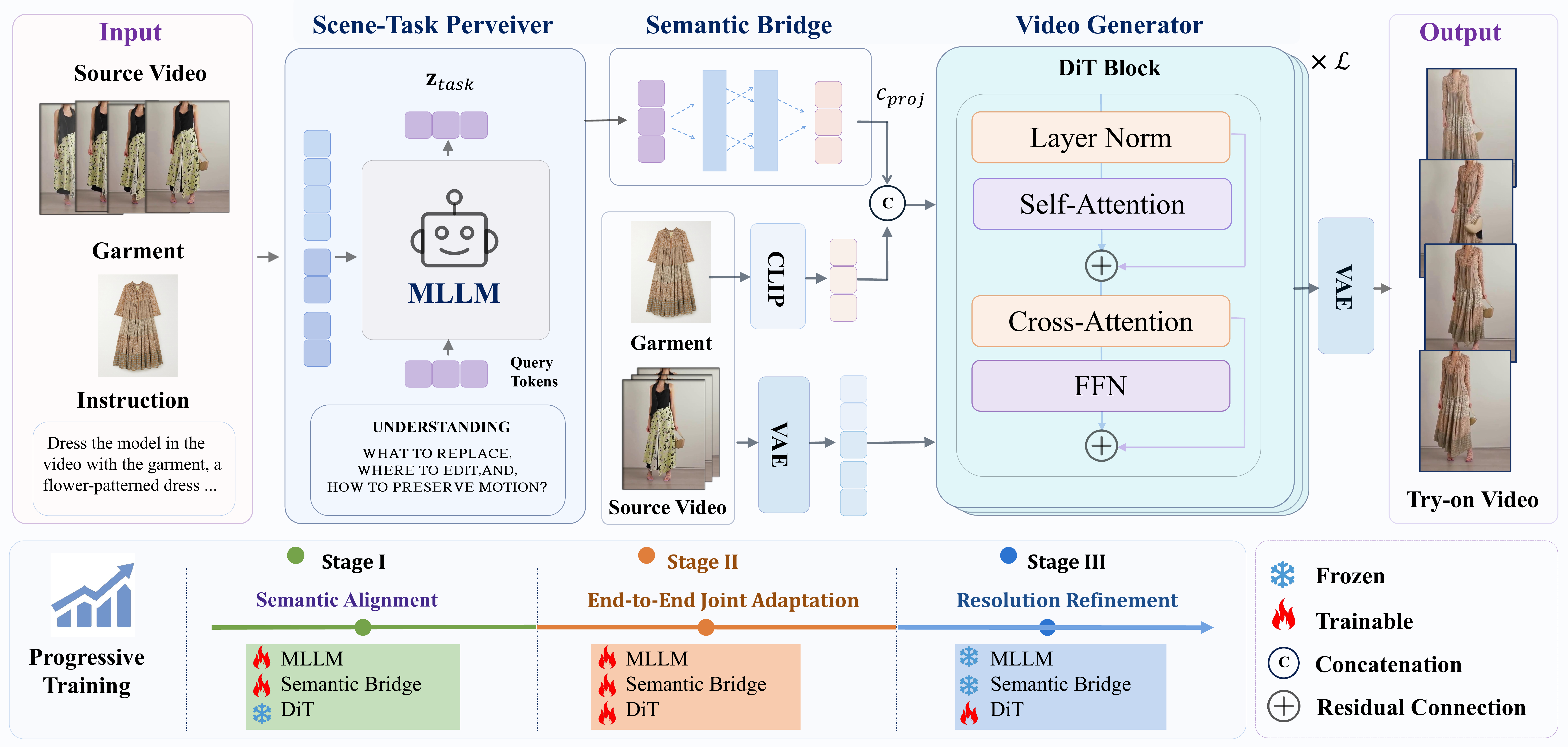}
  \caption{Overview of UniVVT. UniVVT consists of a scene-task perceiver that comprehensively understands the try-on task and extracts compact task-aware representations, which are then aligned to the conditional space of the generative model through a semantic bridge to implicitly guide virtual try-on video generation. A progressive training strategy further enables effective collaboration among components for better adaptation to the VVT task.}
  \label{fig:framework}
\end{figure*}

\section{Methodology}
\subsection{Problem Formulation and Overview}

Given a source video $\mathbf{V}_{\text{in}}$, a target garment image $\mathbf{I}_{\text{gar}}$, and a textual instruction $\mathbf{T}_{\text{inst}}$, video virtual try-on aims to synthesize an output video $\mathbf{V}_{\text{out}}$ that dresses the person in the target garment while preserving the person's identity, original motion, scene context, and temporal coherence.

Unlike inpainting approaches that depend on preprocessing, UniVVT formulates VVT as a video generation task under implicit semantic guidance. Without using explicit masks, poses, or warped garments to prescribe the editing region, UniVVT derives a compact task-aware semantic representation from raw multimodal inputs. As illustrated in Figure~\ref{fig:framework}, UniVVT comprises three components: (1) a multimodal \textbf{Scene-Task Perceiver} that encodes the try-on context, (2) a lightweight \textbf{Semantic Bridge} that maps the resulting representation into the generator's conditioning space, and (3) a \textbf{Video Generation Module} that synthesizes the final video through iterative denoising. By unifying task perception and video generation, UniVVT requires no external modules for explicit geometric encoding at inference.

\subsection{Scene-Task Perceiver}
\label{subsec:perceiver}

The Scene-Task Perceiver transforms the source video, target garment, and textual instruction into a compact representation of the try-on objective. We instantiate it with the pretrained Qwen3-VL-2B-Instruct~\cite{Qwen3-VL}, denoted by $\pi_{\text{mllm}}$, and use its hidden states as multimodal features rather than decoding text.

\paragraph{\textbf{Input tokenization.}}
We uniformly sample key frames from $\mathbf{V}_{\text{in}}$ and convert them into visual tokens using the MLLM's vision encoder. The garment image $\mathbf{I}_{\text{gar}}$ supplies the target appearance, while $\mathbf{T}_{\text{inst}}$ specifies the task. These inputs are organized into a unified token sequence following the native multimodal format of the MLLM.

\paragraph{\textbf{Task-aware latent extraction.}}
We prepend $N$ learnable task queries $\mathbf{Q}_{\text{task}}$ to the multimodal sequence. Through self-attention, these queries aggregate task-relevant information across the instruction, video, and garment tokens. Their final hidden states define the task-aware latent representation
\begin{equation}
\mathbf{z}_{\text{task}} =
\operatorname{Extract}_{\mathbf{Q}}\!\left(
\pi_{\text{mllm}}([\mathbf{Q}_{\text{task}};\mathbf{T}_{\text{inst}};\mathbf{V}_{\text{in}};\mathbf{I}_{\text{gar}}])
\right).
\end{equation}

Unlike masks or pose maps, $\mathbf{z}_{\text{task}}$ jointly captures garment attributes, scene context, body--garment correspondence, and temporal editing intent in a dense semantic representation. It therefore provides a task-specific interface between multimodal understanding and video generation without handcrafted geometric supervision.

\subsection{Semantic Bridge}
\label{subsec:bridge}

The task latent $\mathbf{z}_{\text{task}}$ and the video generator occupy heterogeneous feature spaces, creating both dimensional and distributional mismatches. We therefore introduce a lightweight trainable Semantic Bridge $\mathcal{P}_{\theta}$ that projects $\mathbf{z}_{\text{task}}$ into generator-compatible conditioning tokens:
\begin{equation}
\mathbf{c}_{\text{proj}} = \mathcal{P}_{\theta}(\mathbf{z}_{\text{task}}),
\qquad
\mathbf{c}_{\text{proj}} \in \mathbb{R}^{N \times d_{\text{cond}}}.
\end{equation}

Here, $N$ is the number of task queries and $d_{\text{cond}}$ is the generator's conditioning dimension. Implemented as a compact multi-layer perceptron, $\mathcal{P}_{\theta}$ serves as a semantic adapter rather than learning new generative priors. It translates high-level cues---\emph{what} to transfer, \emph{where} to edit, and \emph{how} to maintain appearance over time---into conditioning signals that modulate the denoising process.

\subsection{Video Generation Module}
\label{subsec:generator}

Our generator $\mathcal{G}_{\phi}$ builds on a pretrained latent video diffusion model with a Diffusion Transformer (DiT) backbone~\cite{wan2025}. It synthesizes a high-fidelity, temporally coherent try-on video through iterative denoising under joint structural, appearance, and semantic conditioning.

\paragraph{\textbf{Latent input construction.}}
A pretrained causal video VAE encodes $\mathbf{V}_{\text{in}}$ and $\mathbf{I}_{\text{gar}}$ into the source-video latent $\mathbf{z}_{\text{video}}$ and garment latent $\mathbf{z}_{\text{gar}}$. At each denoising step, we concatenate $\mathbf{z}_{\text{video}}$ with the noisy latent $\mathbf{x}_t$ along the channel dimension, retaining identity, motion, and background structure. We then append $\mathbf{z}_{\text{gar}}$ along the sequence dimension as a static appearance reference. The resulting latent sequence is denoted by $\mathbf{T}^{\text{in}}_{\text{seq}}$.

\paragraph{\textbf{Condition injection.}}
While $\mathbf{c}_{\text{proj}}$ specifies the desired edit, its compact representation may omit fine garment appearance. We therefore extract a complementary feature $\mathbf{f}_{\text{clip}}$ from $\mathbf{I}_{\text{gar}}$ using a frozen CLIP image encoder~\cite{cherti2023reproducible}. Each DiT block injects both conditions through cross-attention. Specifically, $\mathbf{T}^{\text{in}}_{\text{seq}}$ produces queries $\mathbf{Q}$, whereas the task and CLIP features produce concatenated keys $\mathbf{K}_c=[\mathbf{K}_{\text{task}};\mathbf{K}_{\text{clip}}]$ and values $\mathbf{V}_c=[\mathbf{V}_{\text{task}};\mathbf{V}_{\text{clip}}]$:
\begin{equation}
\operatorname{Attn}(\mathbf{Q},\mathbf{K}_c,\mathbf{V}_c)
=
\operatorname{Softmax}\!\left(
\frac{\mathbf{Q}\mathbf{K}_c^\top}{\sqrt{d}}
\right)\mathbf{V}_c.
\end{equation}

The task condition conveys \emph{what}, \emph{where}, and \emph{how} to edit, while the CLIP feature reinforces garment appearance. Their fusion enables semantically precise editing with faithful garment transfer across video frames.

\subsection{Training Objective}
\label{subsec:objective}

We optimize UniVVT in latent space using conditional flow matching~\cite{esser2024scaling}. Let $\mathbf{x}_0 \sim p_{\mathrm{data}}$ be a clean target-video latent drawn from the latent data distribution $p_{\mathrm{data}}$, $\boldsymbol{\epsilon} \sim \mathcal{N}(\mathbf{0},\mathbf{I})$ be standard Gaussian noise with identity covariance $\mathbf{I}$, and $t \sim \mathcal{U}(0,1)$ be a uniformly sampled time step. The linear probability path and its target velocity are
\begin{equation}
\mathbf{x}_t=(1-t)\mathbf{x}_0+t\boldsymbol{\epsilon},
\qquad
\mathbf{v}_t=\boldsymbol{\epsilon}-\mathbf{x}_0.
\end{equation}

Let $\mathbf{c}=(\mathbf{z}_{\text{video}},\mathbf{z}_{\text{gar}},\mathbf{c}_{\text{proj}},\mathbf{f}_{\text{clip}})$ denote the complete conditioning set, comprising source-video structure, target-garment appearance, and task-aware semantics. We optimize the conditional velocity predictor $\mathcal{G}_{\phi}$ using
\begin{equation}
\mathcal{L} =
\mathbb{E}_{\mathbf{x}_0,\boldsymbol{\epsilon},t}
\left[
\left\|
\mathcal{G}_{\phi}(\mathbf{x}_t,\mathbf{c},t)-\mathbf{v}_t
\right\|_2^2
\right].
\end{equation}

This objective fits the conditional vector field along the data--noise path. At inference, integrating the learned field in reverse transports noise toward a target-video latent while preserving the source motion and scene and faithfully transferring the target garment.

\subsection{Progressive Training Strategy}
\label{subsec:training}

Training UniVVT end to end is challenging because the perceiver and generator originate from distinct pretrained models, while the semantic bridge is initialized from scratch. Their disparate representation spaces and optimization dynamics make naive joint training unstable and yield weak cross-module alignment. We therefore progressively adapt the framework in three stages.

\paragraph{\textbf{Stage 1: Semantic alignment.}}
We first freeze the video generator $\mathcal{G}_{\phi}$, optimize the semantic bridge $\mathcal{P}_{\theta}$, and adapt the MLLM $\pi_{\text{mllm}}$ using Low-Rank Adaptation (LoRA)~\cite{hu2022lora}. With the generator serving as a fixed target interface, the perceiver and bridge learn to produce task-aware conditions compatible with its conditioning space. This stage establishes a reliable semantic interface for stable and effective subsequent joint training.

\paragraph{\textbf{Stage 2: Joint end-to-end adaptation.}}
We then jointly optimize the VVT objective by applying LoRA to both $\pi_{\text{mllm}}$ and $\mathcal{G}_{\phi}$ while fully training $\mathcal{P}_{\theta}$. Training at a base resolution of $512 \times 384$ controls memory cost and improves optimization efficiency. This joint adaptation co-adapts perception, alignment, and generation, enabling faithful garment transfer and temporally coherent synthesis.

\paragraph{\textbf{Stage 3: Flexible-resolution refinement.}}
Finally, we freeze $\pi_{\text{mllm}}$ and $\mathcal{P}_{\theta}$ and continue fine-tuning the LoRA parameters of $\mathcal{G}_{\phi}$ from the Stage 2 checkpoint, using randomly sampled resolutions from 256p to 1024p. This stage improves fidelity and scale robustness without disrupting the semantic alignment learned in earlier stages.

Together, this curriculum decomposes the difficult coupling of heterogeneous pretrained models into semantic alignment, end-to-end co-adaptation, and resolution refinement. It progressively turns MLLM semantics into effective generative control, yielding stable optimization, faithful garment transfer, and flexible-resolution synthesis.

  \section{Experiments}
  \subsection{Datasets}
  We train UniVVT on a mixed image–video corpus consisting of VITON-HD (11,647 pairs at $768\times1024$)~\cite{choi2021viton}, DressCode (48,392 pairs at $768\times1024$)~\cite{morelli2022dresscode}, and ViViD (7,759 pairs at $624\times832$)~\cite{fang2024vivid}. Our training formulation requires source–target pairs of the same subject with consistent pose or motion but different garments, which are hard to collect. We thus train separate DensePose-conditioned inpainters for images and videos to synthesize only source inputs and construct $\langle\textit{source},\textit{garment},\textit{target}\rangle$ triplets. All garment references and ground-truth targets come from original datasets, meaning UniVVT is supervised purely by real images and videos. DensePose is limited to offline data construction and unused during inference. Adopting official train-test splits, we evaluate DressCode and VITON-HD with standard image-level protocols and ViViD-S~\cite{chong2025catv2ton} with the video-level protocol under both paired and unpaired settings.
  \subsection{Evaluation Metrics}
  For image try-on, we use FID~\shortcite{Seitzer2020FID} and KID~\shortcite{binkowski2018demystifying} to measure distribution-level realism. Under paired settings, we additionally report SSIM~\shortcite{wang2004image} and LPIPS~\shortcite{zhang2018unreasonable} for structural and perceptual similarity, respectively. For video try-on, we report VFID computed with I3D~\shortcite{carreira2017quo} and ResNeXt~\shortcite{xie2017aggregated} backbones, denoted as VFID$_\mathrm{I}$ and VFID$_\mathrm{R}$. Paired video evaluation further includes frame-level SSIM and LPIPS.
  
  \subsection{Implementation Details}
  We conduct all experiments on eight NVIDIA A100 GPUs. The scene-task perceiver and video generator are initialized from Qwen3-VL-2B-Instruct~\shortcite{Qwen3-VL} and Wan2.1-Fun-Control~\shortcite{wan2025}, respectively, with $N=512$ task queries. We optimize the model using AdamW~\shortcite{loshchilov2017decoupled} with a constant learning rate of $1\times10^{-4}$, weight decay of 0.01, LoRA rank of 32, and a per-GPU batch size of 1. Each training stage runs for 30k steps. Stages 1 and 2 use a resolution of $512\times384$, whereas Stage 3 performs multi-resolution training. At inference, we adopt the Euler scheduler~\shortcite{karras2022elucidating} with 28 sampling steps and a fixed random seed of 42. For fair comparison, all quantitative results are reported at $512\times384$, although UniVVT supports higher-resolution synthesis.
  
  \begin{table*}[htbp]
    \centering
    \small
    \setlength{\tabcolsep}{1.5mm}
    \renewcommand{\arraystretch}{0.8}
    \begin{tabular}{@{}l c cccc cc@{}}
      \toprule
      \multirow{2}{*}{\textbf{Method}} &
      \multirow{2}{*}{\textbf{Paradigm}} &
      \multicolumn{4}{c}{\textbf{Paired}} &
      \multicolumn{2}{c}{\textbf{Unpaired}} \\
      \cmidrule(lr){3-6}\cmidrule(l){7-8}
      & & \textbf{VFID$_\mathrm{I}$} $\downarrow$
      & \textbf{VFID$_\mathrm{R}$} $\downarrow$
      & \textbf{SSIM} $\uparrow$
      & \textbf{LPIPS} $\downarrow$
      & \textbf{VFID$_\mathrm{I}$} $\downarrow$
      & \textbf{VFID$_\mathrm{R}$} $\downarrow$ \\
      \midrule
      StableVITON + AM~\shortcite{kim2024stableviton} & Pre+Inp+AM & 34.2446 & 0.7735 & 0.8019 & 0.1338 & 36.8985 & 0.9064 \\
      OOTDiffusion + AM~\shortcite{xu2025ootdiffusion} & Pre+Inp+AM & 29.5253 & 3.9372 & 0.8087 & 0.1232 & 35.3170 & 5.7078 \\
      IDM-VTON + AM~\shortcite{choi2024improving} & Pre+Inp+AM & 20.0812 & 0.3674 & 0.8227 & 0.1163 & 25.4972 & 0.7167 \\
      ViViD~\shortcite{fang2024vivid} & Pre+Inp & 17.2924 & 0.6209 & 0.8029 & 0.1221 & 21.8032 & 0.8212 \\
      CatV$^2$TON~\shortcite{chong2025catv2ton} & Pre+Inp & 13.5962 & 0.2963 & 0.8727 & 0.0639 & 19.5131 & 0.5283 \\
      OIE~\shortcite{pan2025once} & Pre+Inp & 9.3983 & -- & 0.8466 & 0.0774 & 17.0831 & -- \\
      DreamVVT~\shortcite{zuo2025dreamvvt} & Pre+Inp & 11.0180 & 0.2549 & 0.8737 & 0.0619 & 16.9468 & 0.4285 \\
      MagicTryOn~\shortcite{li2025magictryon} & Pre+Inp & \underline{8.4030} & \underline{0.2346} & \textbf{0.9011} & \underline{0.0602} & \underline{14.7174} & \underline{0.3200} \\
      \midrule
      \rowcolor{red!5}
      \textbf{Ours} & \textbf{End2End} & \textbf{8.3623} & \textbf{0.1934} & \underline{0.8922} & \textbf{0.0456} & \textbf{12.3640} & \textbf{0.1876} \\
      \bottomrule
    \end{tabular}
    \caption{Quantitative results on ViViD-S under paired and unpaired settings. Arrows indicate the preferred direction. Best and second-best results are bold and underlined, respectively. Pre, Inp, and AM denote preprocessing, inpainting, and animation.}
    \label{tab:exp_vivid}
  \end{table*}
  
    \begin{table}[htbp]
      \centering
      \small
      \renewcommand{\arraystretch}{0.8}
      \setlength{\tabcolsep}{0.8mm}
      \begin{tabular}{@{}l cccc cc@{}}
        \toprule
        \multirow{2}{*}{\textbf{Method}} &
        \multicolumn{4}{c}{\textbf{Paired}} &
        \multicolumn{2}{c}{\textbf{Unpaired}} \\
        \cmidrule(lr){2-5}\cmidrule(l){6-7}
        & \textbf{FID} $\downarrow$
        & \textbf{KID} $\downarrow$
        & \textbf{SSIM} $\uparrow$
        & \textbf{LPIPS} $\downarrow$
        & \textbf{FID} $\downarrow$
        & \textbf{KID} $\downarrow$ \\
        \midrule
        GP-VTON & 9.927 & 4.610 & 0.7711 & 0.1801 & 12.792 & 6.627 \\
        LaDI-VTON & 9.555 & 4.683 & 0.7656 & 0.2366 & 10.676 & 5.787 \\
        IDM-VTON & 6.821 & 2.924 & 0.8797 & 0.0563 & 9.546 & 4.320 \\
        OOTDiffusion & 4.610 & 0.955 & 0.8854 & 0.0533 & 12.567 & 6.627 \\
        CatVTON & 3.992 & 0.818 & 0.8922 & 0.0455 & 6.137 & 1.549 \\
        CatV$^2$TON & 5.722 & 2.338 & \textbf{0.9222} & \underline{0.0367} & 8.627 & 3.838 \\
        MagicTryOn & \underline{3.356} & \underline{0.6851} & 0.9032 & 0.04215 & \underline{5.314} & \underline{1.339} \\
        \midrule
        \rowcolor{red!5}
        \textbf{Ours} & \textbf{2.734} & \textbf{0.4154} & \underline{0.9130} & \textbf{0.0349} & \textbf{4.900} & \textbf{0.960} \\
        \bottomrule
      \end{tabular}
      \caption{Quantitative results on DressCode under paired and unpaired settings. Arrows indicate the preferred direction; best and second-best results are bold and underlined.}
      \label{tab:exp_dresscode}
    \end{table}
    \subsection{Quantitative Experiments}
  \paragraph{Performance.} Table~\ref{tab:exp_vivid} reports quantitative results on the ViViD-S video try-on benchmark. UniVVT delivers the best overall performance on most metrics: it achieves the lowest VFID$_\mathrm{I}$ and VFID$_\mathrm{R}$ under both paired and unpaired settings, indicating improved video realism and closer alignment with the real-data distribution. Under paired evaluation, UniVVT obtains the best LPIPS and a competitive SSIM, demonstrating strong perceptual quality and structural consistency. Tables~\ref{tab:exp_dresscode} and~\ref{tab:exp_viton_hd} present results on the DressCode and VITON-HD image try-on benchmarks. UniVVT consistently achieves the best or second-best performance across FID, KID, SSIM, and LPIPS, demonstrating consistent effectiveness across video and image try-on. Notably, the competing methods rely on explicit preprocessing and inpainting-based generation, whereas UniVVT performs end-to-end conditional synthesis without such modules at inference. These results validate the effectiveness of unifying task understanding and generation while eliminating fragile upstream dependencies.
    \begin{table}[htbp]
      \centering
      \small
      \renewcommand{\arraystretch}{0.8}
      \setlength{\tabcolsep}{0.8mm}
      \begin{tabular}{@{}l cccc cc@{}}
        \toprule
        \multirow{2}{*}{\textbf{Method}} &
        \multicolumn{4}{c}{\textbf{Paired}} &
        \multicolumn{2}{c}{\textbf{Unpaired}} \\
        \cmidrule(lr){2-5}\cmidrule(l){6-7}
        & \textbf{FID} $\downarrow$
        & \textbf{KID} $\downarrow$
        & \textbf{SSIM} $\uparrow$
        & \textbf{LPIPS} $\downarrow$
        & \textbf{FID} $\downarrow$
        & \textbf{KID} $\downarrow$ \\
        \midrule
        GP-VTON & 8.726 & 3.944 & 0.8701 & 0.0585 & 11.844 & 4.310 \\
        LaDI-VTON & 11.386 & 7.248 & 0.8603 & 0.0733 & 14.648 & 8.754 \\
        IDM-VTON & 6.338 & 1.322 & 0.8806 & 0.0789 & 9.611 & 1.639 \\
        OOTDiffusion & 9.305 & 4.086 & 0.8187 & 0.0876 & 12.408 & 4.689 \\
        CatVTON & 6.139 & 0.964 & 0.8691 & 0.0973 & \underline{9.143} & 1.267 \\
        CatV$^2$TON & 8.095 & 2.245 & \textbf{0.8902} & 0.0572 & 11.222 & 2.986 \\
        MagicTryOn & \underline{5.689} & \underline{0.710} & 0.8804 & \underline{0.0526} & 9.203 & \underline{1.217} \\
        \midrule
        \rowcolor{red!5}
        \textbf{Ours} & \textbf{5.348} & \textbf{0.3667} & \underline{0.8821} & \textbf{0.0512} & \textbf{9.014} & \textbf{1.130} \\
        \bottomrule
      \end{tabular}
      \caption{Quantitative results on VITON-HD under paired and unpaired settings. Arrows indicate the preferred direction; best and second-best results are bold and underlined.}
      \label{tab:exp_viton_hd}
    \end{table}

\begin{figure}[tb] 
    \centering
    \includegraphics[width=\linewidth]{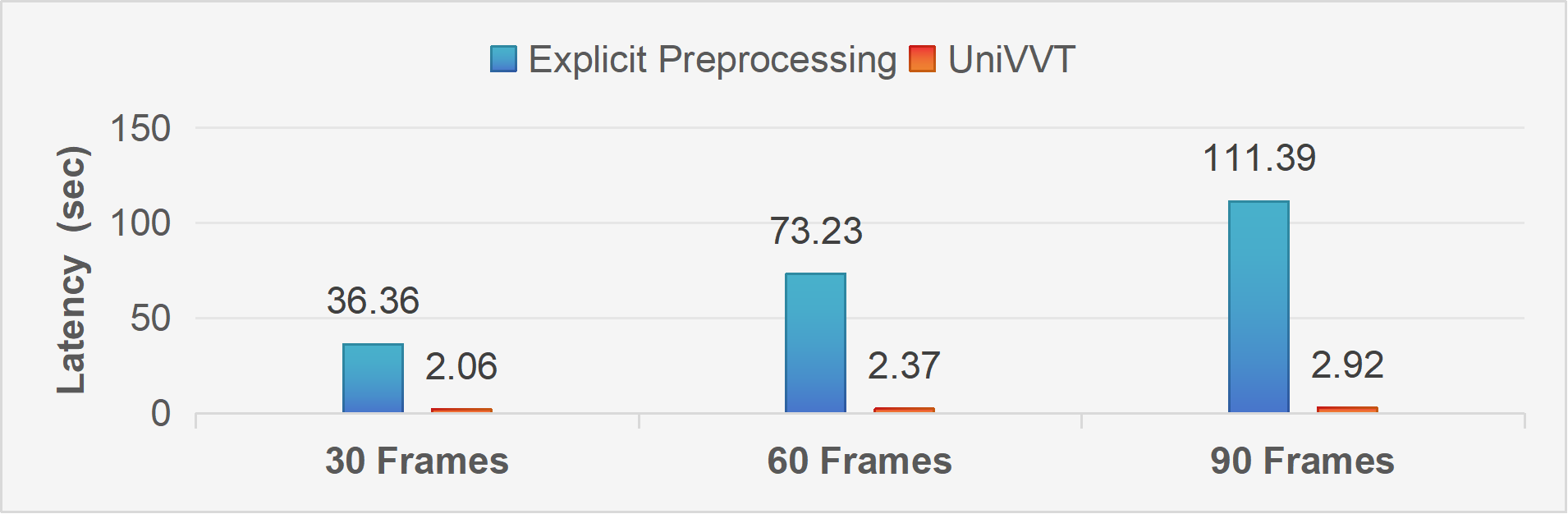}
    \caption{Conditioning latency versus video length. Explicit geometric preprocessing scales nearly linearly, while UniVVT's implicit task encoding grows only marginally.}
    \label{fig:latency_comparison}
    \vspace{-1em}
  \end{figure}

  \paragraph{Latency analysis.} Figure~\ref{fig:latency_comparison} compares the conditioning overhead of frame-wise geometric preprocessing, including mask and DensePose extraction, with UniVVT's implicit task encoding. As the video length increases from 30 to 90 frames, explicit preprocessing grows nearly linearly from 36.36\,s to 111.39\,s, whereas UniVVT increases only from 2.06\,s to 2.92\,s. UniVVT therefore delivers a \textbf{17.7--38.1$\times$} speedup, with the advantage widening as videos become longer; for 90 frames, it removes 108.47\,s of preprocessing overhead. This gain is architectural rather than merely implementational: replacing dense frame-wise geometry extraction with a compact semantic representation fundamentally improves scalability while preserving end-to-end task awareness.
  \begin{figure*}
    \centering
      \includegraphics[width=\textwidth]{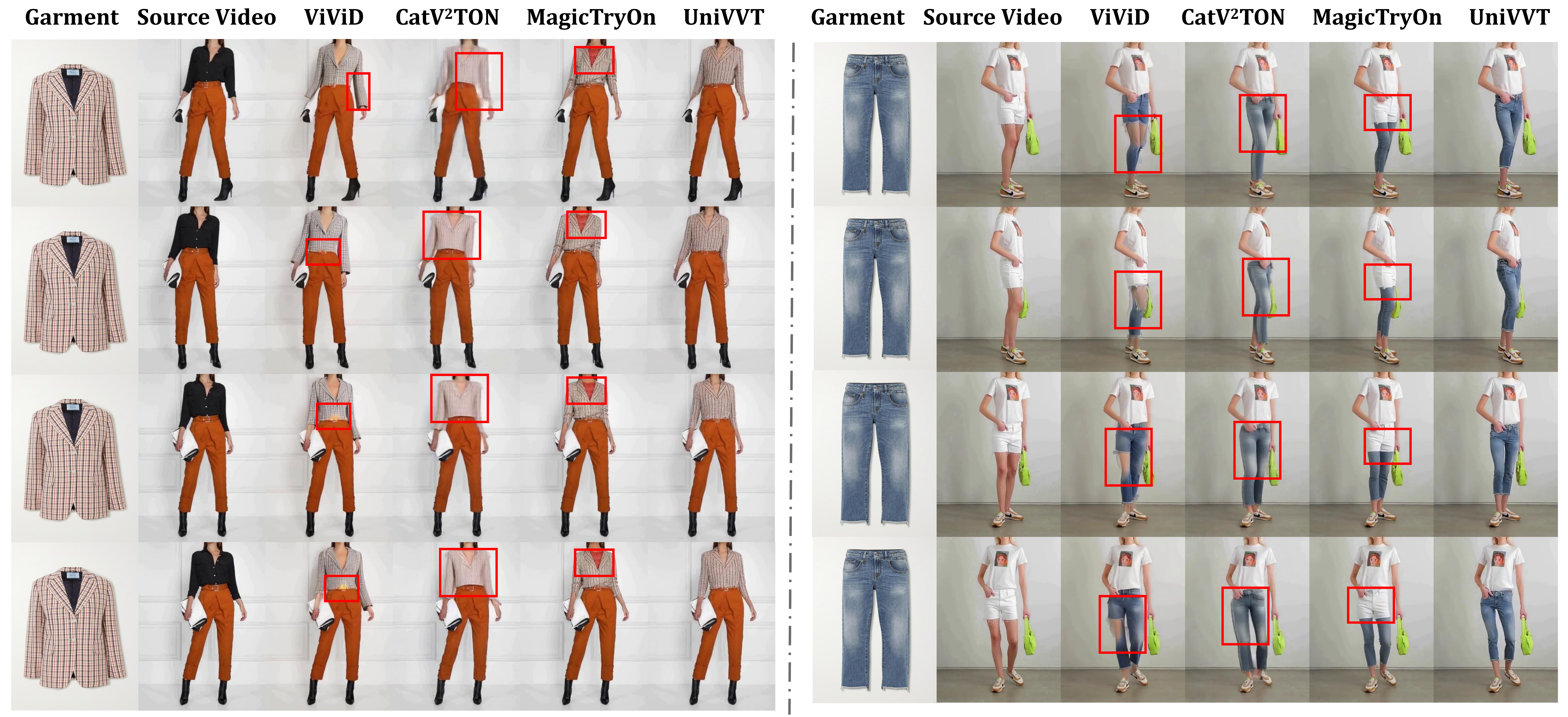}
      \caption{Qualitative comparison on the ViViD-S benchmark under unpaired settings. Compared with baseline methods, UniVVT produces videos with higher visual fidelity and better garment consistency. Please zoom in for better visualization.}
      \label{fig:compare_vivid}
    \end{figure*}

  \subsection{Qualitative Experiments}
\begin{figure}[tb] 
    \centering
        \includegraphics[width=\linewidth]{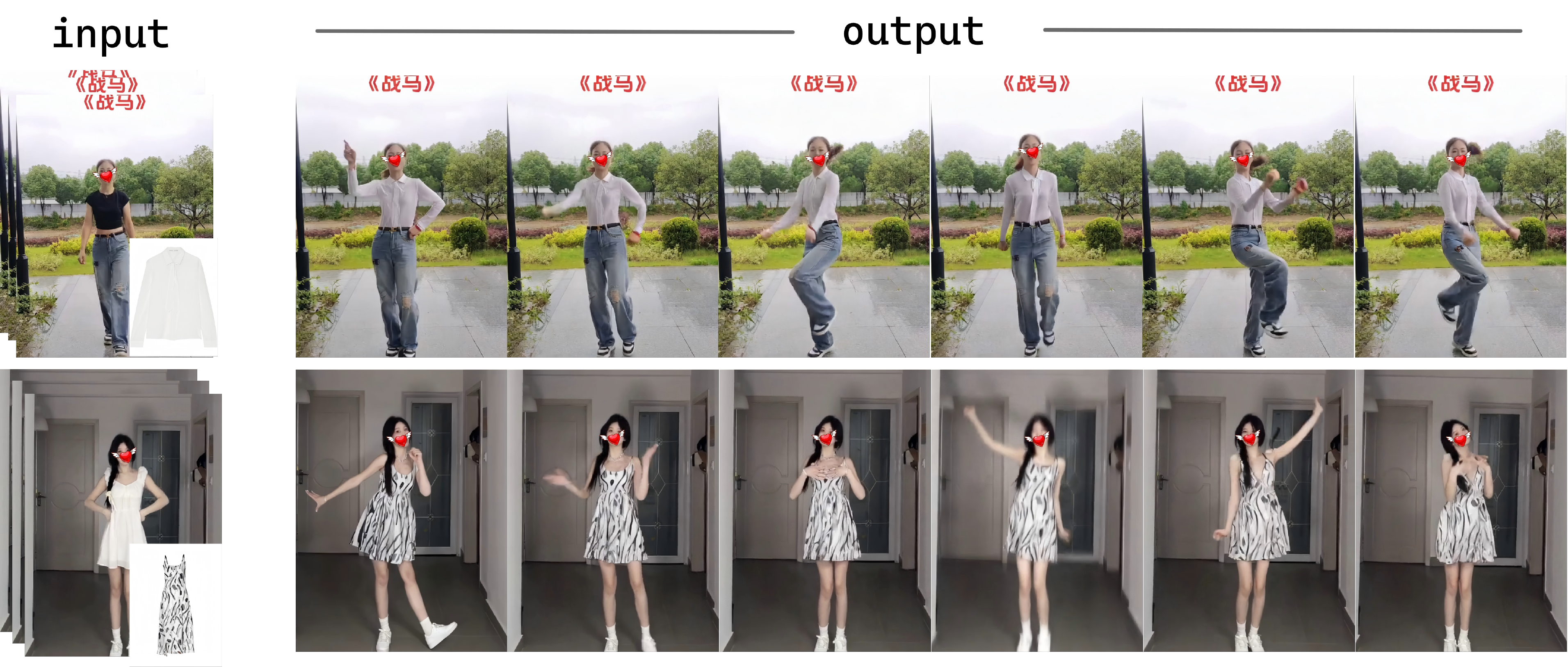}
        \caption{In-the-wild try-on results of UniVVT on dance videos with complex motion.} 
        \label{fig:in-the-wild}
    \end{figure}

\begin{figure}[h] 
    \centering
        \includegraphics[width=\linewidth]{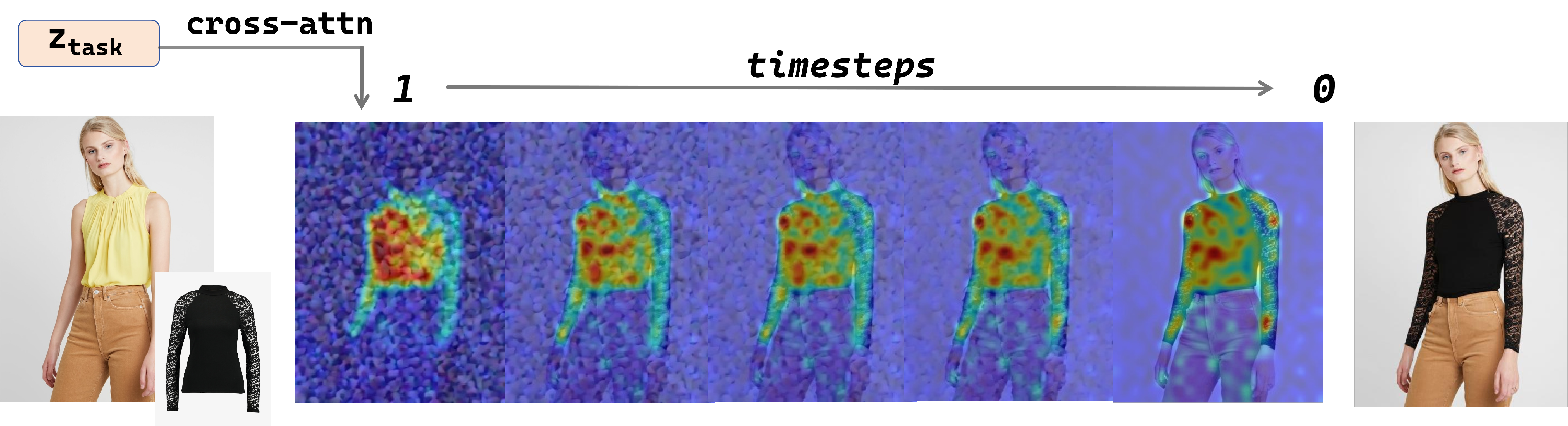}
        \caption{Cross-attention of $\mathbf{z}_{\text{task}}$ over denoised latents. Attention focuses on the editable garment region while remaining weak on identity and background.} 
        \label{fig:cross-attn-vis}
    \end{figure}

\begin{figure}[h]
    \centering
    \includegraphics[width=\linewidth]{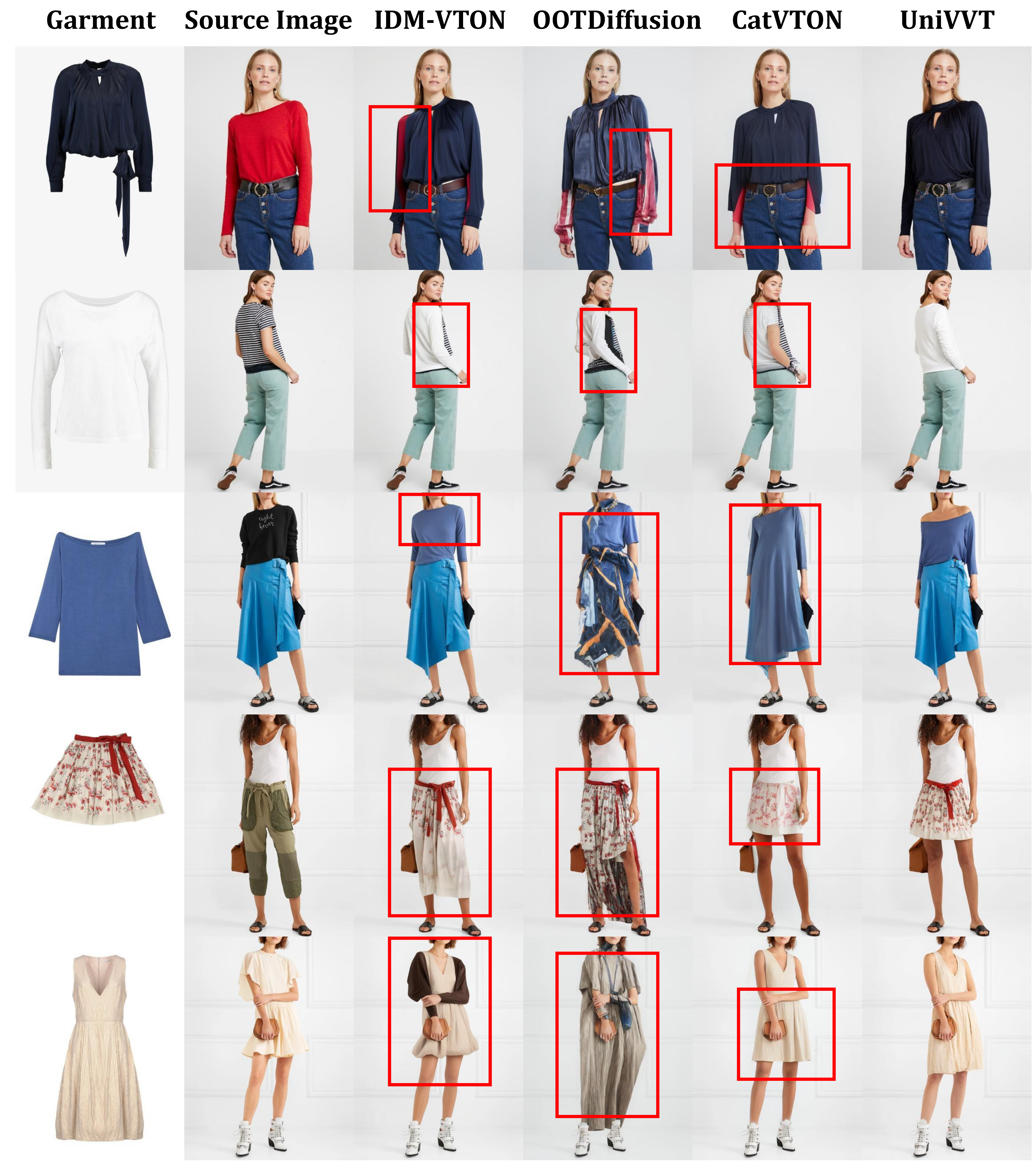}
    \caption{Qualitative comparison on VITON-HD (top two rows) and DressCode (bottom three rows). UniVVT better preserves garment appearance, pose, and background.}
      \label{fig:compare_dresscode}
  \end{figure}

  \paragraph{Visual comparision.} Figure~\ref{fig:compare_vivid} provides visual comparisons between UniVVT and recent video virtual try-on methods under the unpaired setting on ViViD-S. UniVVT generates outputs with sharper garment textures and stronger temporal consistency. By contrast, ViViD and CatV$^2$TON frequently suffer from blurred textures, missing details, and boundary artifacts. Although MagicTryOn alleviates some defects via fine-grained feature modeling, incomplete garment replacement still occurs when mask estimation is inaccurate. Figure~\ref{fig:compare_dresscode} further evaluates image try-on performance for diverse garment categories on DressCode and VITON-HD. UniVVT accurately locates editing regions, preserves human poses and background content, and faithfully transfers the color, texture, and style of reference garments. In comparison, IDM-VTON and CatVTON struggle to retain garment styles, while OOTDiffusion tends to introduce structural artifacts. These observations demonstrate that UniVVT avoids reliance on fragile preprocessing at inference and mitigates try-on failures caused by upstream error propagation, highlighting the benefit of end-to-end optimization.
  \paragraph{In-the-wild video try-on.}
  Figure~\ref{fig:in-the-wild} presents UniVVT try-on results on in-the-wild TikTok dance videos. These complex dynamic scenes are out-of-distribution (OOD) relative to our training data, with no corresponding samples used during optimization. UniVVT nevertheless adapts well to such unseen real-world settings, demonstrating strong generalization and robustness.
  \paragraph{Interpretation via attention visualization.}
  Why can UniVVT localize the spatio-temporal edit region without an explicit mask prior? Unlike conventional pipelines that hard-constrain synthesis with garment masks, UniVVT conditions the generator with a compact task representation. We visualize the cross-attention of $\mathbf{z}_{\text{task}}$ over the denoised latents during generation (Figure~\ref{fig:cross-attn-vis}). For a long-sleeved target, attention concentrates on the upper body and remains high along the arms across frames, while responses on the face and background stay comparatively weak. This indicates that $\mathbf{z}_{\text{task}}$ jointly encodes \emph{what} to transfer, \emph{where} to edit, and \emph{how} to maintain appearance over time, acting as an implicit semantic guide that enables precise try-on without mask supervision. 

  \begin{figure}[t]
    \centering
    \includegraphics[width=\linewidth]{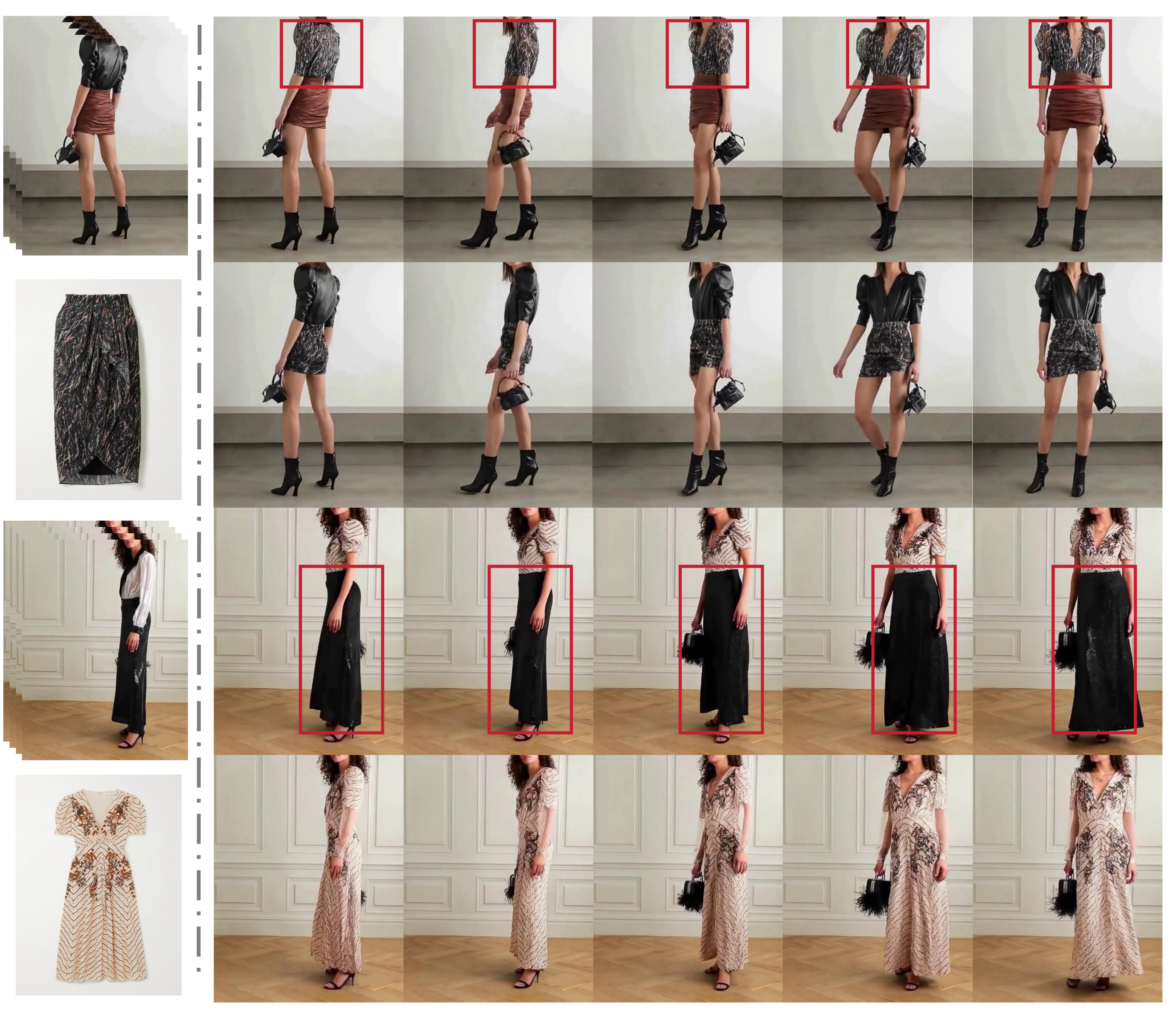}
    \caption{Qualitative ablation of the scene-task perceiver. Left: source video and reference garment; right: w/o STP (rows 1 and 3) vs.\ UniVVT (rows 2 and 4).}
      \label{fig:ablation_mllm}
  \end{figure}
  \begin{table}[htbp]
    \centering
    \footnotesize
    \renewcommand{\arraystretch}{0.8}
    \setlength{\tabcolsep}{0.35mm}
    \begin{tabular}{@{}lcccccc@{}}
      \toprule
      \multirow{2}{*}{\textbf{Method}} &
      \multicolumn{4}{c}{\textbf{Paired}} &
      \multicolumn{2}{c}{\textbf{Unpaired}} \\
      \cmidrule(lr){2-5}\cmidrule(l){6-7}
      & \makecell{\textbf{VFID$_\mathrm{I}$}$\downarrow$}
      & \makecell{\textbf{VFID$_\mathrm{R}$}$\downarrow$}
      & \makecell{\textbf{SSIM}$\uparrow$}
      & \makecell{\textbf{LPIPS}$\downarrow$}
      & \makecell{\textbf{VFID$_\mathrm{I}$}$\downarrow$}
      & \makecell{\textbf{VFID$_\mathrm{R}$}$\downarrow$} \\
      \midrule
      \textbf{UniVVT} & \textbf{8.3623} & \textbf{0.1934} & \textbf{0.8922} & \textbf{0.0456} & \textbf{12.3640} & \textbf{0.1876} \\
      w/o STP & 11.7308 & 0.2282 & 0.8509 & 0.0752 & 14.2424 & 0.4424 \\
      \bottomrule
    \end{tabular}
    \caption{Ablation study of the STP module on ViViD-S. Arrows indicate the preferred direction; best results are bold.}
    \label{tab:ablation_mllm}
  \end{table}
  \subsection{Ablation Studies}
  \paragraph{Necessity of the scene-task perceiver.}
  The scene-task perceiver (STP) maps multimodal inputs into a compact task latent $\mathbf{z}_{\text{task}}$ that captures overall try-on intent. To assess its necessity, we replace STP with a text-only encoder, yielding the variant \textbf{w/o STP}. As shown in Table~\ref{tab:ablation_mllm}, this change degrades all metrics under both paired and unpaired settings. Figure~\ref{fig:ablation_mllm} further shows that, without STP, the model loses holistic task understanding and fails to localize valid edit regions. For example, a short skirt is incorrectly placed on the upper body (row 1), and only the upper half of a target dress is transferred while the lower body remains unchanged (row 3). These results confirm that STP is critical for macroscopic task understanding and intent control in virtual try-on.

  \begin{table}[htbp]
    \centering
    \small
    \renewcommand{\arraystretch}{0.8}
    \setlength{\tabcolsep}{1.0mm}
    \begin{tabular}{@{}ll cccc@{}}
      \toprule
      \multirow{2}{*}{\textbf{Phase}} &
      \multirow{2}{*}{\textbf{Resolution}} &
      \multicolumn{4}{c}{\textbf{Paired}} \\
      \cmidrule(l){3-6}
      & & \textbf{VFID$_\mathrm{I}$} $\downarrow$
      & \textbf{VFID$_\mathrm{R}$} $\downarrow$
      & \textbf{SSIM} $\uparrow$
      & \textbf{LPIPS} $\downarrow$ \\
      \midrule
      Stage 1 & $512\times384$ & 21.5989 & 1.9006 & 0.6723 & 0.1873 \\
      \midrule
      \multirow{3}{*}{Stage 2}
        & $256\times192$ & 8.7705 & 0.1804 & 0.8668 & 0.0476 \\
        & $512\times384$ & 8.3050 & 0.1981 & 0.8923 & 0.0454 \\
        & $832\times624$ & 8.5018 & 0.1702 & 0.8760 & 0.0597 \\
      \midrule
      \multirow{3}{*}{Stage 3}
        & $256\times192$ & 8.8597 & \textbf{0.1778} & \textbf{0.8896} & \textbf{0.0382} \\
        & $512\times384$ & 8.3623 & 0.1934 & 0.8922 & 0.0456 \\
        & $832\times624$ & \textbf{8.2969} & 0.1774 & \textbf{0.9062} & \textbf{0.0586} \\
      \bottomrule
    \end{tabular}
    \caption{Stage-wise progressive training results on ViViD-S (paired). Arrows indicate the preferred direction; bold marks Stage~3 gains at non-default resolutions.}
    \label{tab:ablation_progress_training}
  \end{table}
  
  \paragraph{Effect of progressive training.}
  Table~\ref{tab:ablation_progress_training} reports quantitative scores under the paired setting after each training stage. Stage~1 aligns the MLLM and generator conditioning spaces with a frozen DiT backbone; the large residual gap to real data shows that semantic alignment alone is insufficient. Stage~2 jointly adapts all components at $512\times384$ and yields the main gains in distributional realism and garment fidelity. Stage~3 then refines the generator under multi-resolution sampling, further improving SSIM and LPIPS at $256\times192$ and $832\times624$ while preserving competitive quality at the base resolution. Overall, the curriculum progressively converts cross-module alignment into stable VVT performance and resolution flexibility.

\section{Conclusion}
We present UniVVT, a unified end-to-end framework that reformulates video virtual try-on as semantically conditioned video generation rather than mask-conditioned inpainting. An MLLM-based scene-task perceiver and a lightweight semantic bridge convert the source video, target garment, and instruction into compact implicit guidance of \emph{what}, \emph{where}, and \emph{how} to edit, removing fragile mask, pose, and warping modules at inference, while a three-stage progressive training strategy stably couples these heterogeneous pretrained components for reliable try-on synthesis. Extensive experiments on video and image benchmarks show that UniVVT delivers strong garment fidelity and temporal consistency across diverse garments and motion without sacrificing identity or scene context, establishing end-to-end joint optimization with implicit semantic guidance as a practical alternative to multi-stage pipelines limited by disjoint component-wise training and brittle geometric preprocessing.

\bibliography{univvt}
\newpage

\appendix
This appendix complements the main paper with further analyses and implementation details. We first describe the construction of training triplets, and then provide additional ablation studies and additional quantitative and qualitative evaluations, followed by a discussion of the current limitations of UniVVT.

\section{Training-Triplet Construction}
\label{appendix:data_preparation}
\begin{figure*}[tb]
  \centering
  \includegraphics[width=0.95\textwidth]{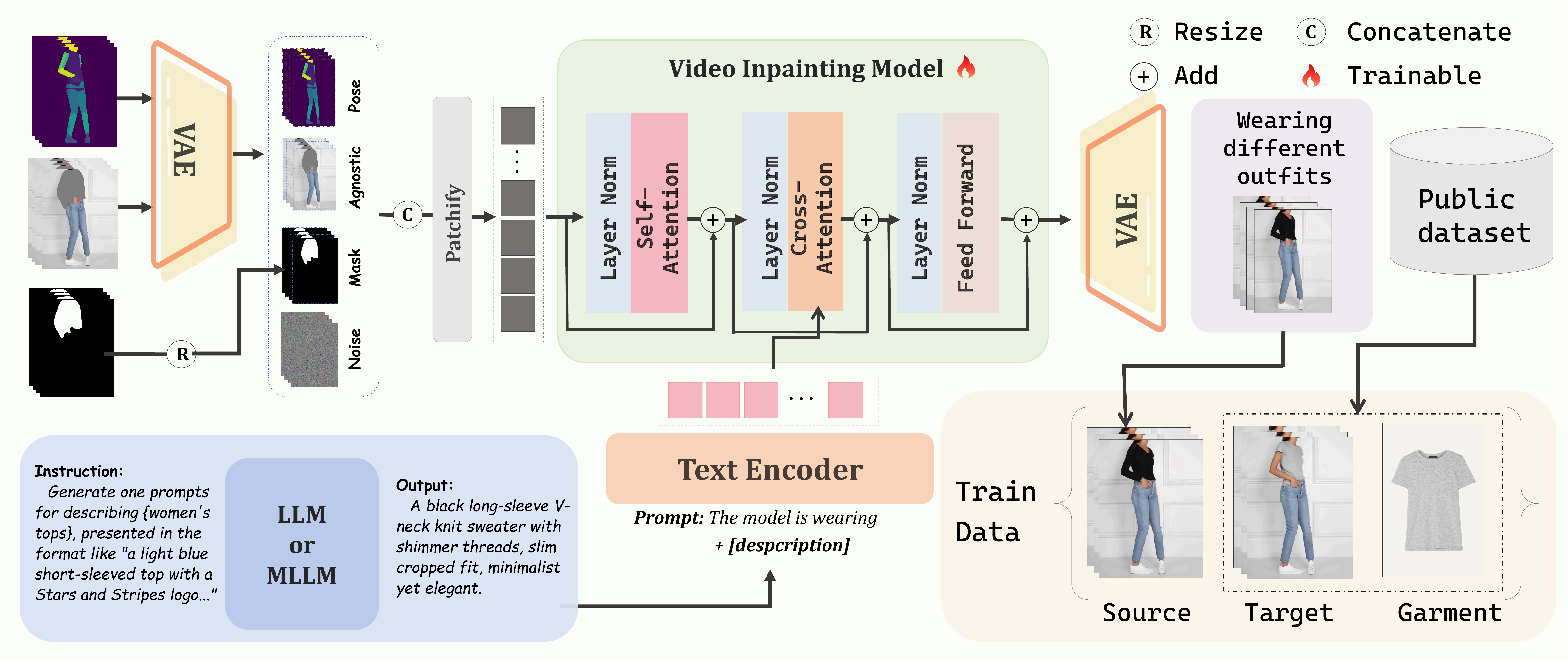}
\caption{Construction of aligned training triplets using synthetic source inputs and real-target supervision. Starting from a real garment--target pair $(\mathbf{I}_{\text{gar}}, \mathbf{V}_{\text{out}})$, we synthesize $\mathbf{V}_{\text{in}}$ by replacing the clothing in $\mathbf{V}_{\text{out}}$ with an alternative garment specified by an LLM-generated description. Only $\mathbf{V}_{\text{in}}$ is synthesized; $\mathbf{I}_{\text{gar}}$ and $\mathbf{V}_{\text{out}}$ are retained as real samples from the original dataset.}
  \label{fig:data_preparation}
\end{figure*}
\paragraph{Motivation.}
Public virtual try-on datasets provide real garment--target pairs $(\mathbf{I}_{\text{gar}}, \mathbf{V}_{\text{out}})$, but UniVVT requires aligned triplets $\langle \mathbf{V}_{\text{in}}, \mathbf{I}_{\text{gar}}, \mathbf{V}_{\text{out}} \rangle$. Ideally, $\mathbf{V}_{\text{in}}$ and $\mathbf{V}_{\text{out}}$ should depict the same person under identical pose, motion, viewpoint, and scene conditions, differing only in the worn garment. Capturing such pairs would require reproducing an entire image or video sequence after a clothing change, which is impractical at scale. Naively pairing unrelated samples would instead entangle garment transfer with changes in identity, motion, camera, and background, weakening the supervision signal.

\paragraph{Triplet construction.}
We address this data gap by synthesizing the source input rather than the supervision target. As shown in Figure~\ref{fig:data_preparation}, we begin with a real pair $(\mathbf{I}_{\text{gar}}, \mathbf{V}_{\text{out}})$ and sample an alternative garment description that is distinct from the target garment. An inpainting model then replaces the clothing in $\mathbf{V}_{\text{out}}$ according to this description, producing a synthetic source $\mathbf{V}_{\text{in}}$. During synthesis, garment masks localize the editable region, while DensePose conditioning preserves body configuration and frame-to-frame motion. Because identity, pose, camera, and background are inherited from the same real target, the resulting triplet isolates the garment change while minimizing unrelated variation.

\paragraph{Real-target supervision.}
Importantly, synthesis is restricted to the conditioning side of the training pair. The generated $\mathbf{V}_{\text{in}}$ serves only as an input, whereas $\mathbf{I}_{\text{gar}}$ and the ground-truth $\mathbf{V}_{\text{out}}$ are retained directly from the original dataset. Consequently, every reconstruction and diffusion target is derived from real imagery: UniVVT is optimized to map a synthetic source back to a real target and is never trained to reproduce an inpainted output. The inpainting models therefore act exclusively as offline data-generation tools, rather than teachers, and no teacher--student distillation is involved. This separation enables synthetic editing to resolve the scarcity of aligned triplets without introducing synthetic targets into the supervision.

\paragraph{Synthesis and quality control.}
We use FLUX.1-dev~\cite{labs2025flux} to construct image triplets and Wan2.1-Fun-1.3B-Control~\cite{wan2025} to construct video triplets. DensePose maps and garment masks are used only during this offline synthesis procedure; they are not exposed to UniVVT as inputs during either training or inference. Finally, we manually inspect every generated source and discard samples with identity drift, inconsistent motion, mask leakage, incomplete garment replacement, or conspicuous inpainting artifacts. This filtering step ensures that the retained triplets preserve the intended correspondence between source content and real target supervision.

\section{Additional Ablation Experiments} 
\begin{table*}[tb]
  \centering
  \setlength{\tabcolsep}{2mm}
  \begin{tabular}{l|cccc|cc}
    \toprule
    \multirow{2}{*}{{Methods}} &
    \multicolumn{4}{c|}{{Paired}} &
    \multicolumn{2}{c}{{Unpaired}} \\
    \hhline{~|----|--}
    & {VFID$_\mathrm{I}\downarrow$} & {VFID$_\mathrm{R}\downarrow$} & {SSIM$\uparrow$} & {LPIPS$\downarrow$} & {VFID$_\mathrm{I}\downarrow$} & {VFID$_\mathrm{R}\downarrow$} \\
    \hhline{-------}
    w/o $\mathbf{z}_{\text{task}}$ & 11.7308 & 0.2282 & 0.8509 & 0.0752 & 14.2424 & 0.4424 \\
    w/o $\mathbf{f}_{\text{clip}}$ & 8.9210 & 0.1996 & 0.8843 & 0.0500 & 12.5608 & 0.1985 \\
    \hhline{-------}
    \textbf{UniVVT} & \textbf{8.3623} & \textbf{0.1934} & \textbf{0.8922} & \textbf{0.0456} & \textbf{12.3640} & \textbf{0.1876} \\
    \bottomrule
  \end{tabular}
  \caption{Quantitative ablation of task-aware semantic guidance and CLIP garment conditioning on the ViViD-S video benchmark.}
  \label{tab:ablation_task_clip}
\end{table*}
\paragraph{Complementarity of task semantics and garment appearance.}
Each DiT block receives two semantic conditions: the projected task representation $\mathbf{c}_{\text{proj}}=\mathcal{P}_{\theta}(\mathbf{z}_{\text{task}})$, which is inferred jointly from the source video, target garment, and instruction, and the garment feature $\mathbf{f}_{\text{clip}}$, extracted independently by a frozen CLIP image encoder. Although both conditions are fused within the same cross-attention layer, they are intended to serve distinct roles. We isolate their contributions by retraining two variants, each removing one conditioning branch while preserving the architecture and optimization protocol otherwise. Table~\ref{tab:ablation_task_clip} shows that removing $\mathbf{z}_{\text{task}}$ degrades all metrics substantially: paired VFID$_\mathrm{I}$ increases from 8.3623 to 11.7308, LPIPS rises from 0.0456 to 0.0752, and unpaired VFID$_\mathrm{R}$ more than doubles from 0.1876 to 0.4424. In contrast, removing $\mathbf{f}_{\text{clip}}$ produces consistently smaller, yet non-negligible, drops. These results identify the task-aware pathway as the principal source of semantic control. Unlike the source-independent CLIP descriptor, $\mathbf{z}_{\text{task}}$ integrates the complete try-on context and can therefore coordinate \emph{what} garment to transfer with \emph{where} and \emph{how} to apply the edit throughout motion. The localized responses visualized in Figure~\ref{fig:cross-attn-vis} of the main paper further support this interpretation. Meanwhile, the consistent gains from $\mathbf{f}_{\text{clip}}$ confirm its complementary role in preserving garment-specific category, style, and appearance cues that may be compressed by the task representation. Their combination consequently achieves the best performance across all paired and unpaired metrics.
\begin{table*}[htbp]
  \centering
  \setlength{\tabcolsep}{2mm}
  \begin{tabular}{c|cccc|cc}
    \toprule
    \multirow{2}{*}{{LoRA MLLM}} &
    \multicolumn{4}{c|}{{Paired}} &
    \multicolumn{2}{c}{{Unpaired}} \\
    \hhline{~|----|--}
    & {VFID$_\mathrm{I}\downarrow$}
    & {VFID$_\mathrm{R}\downarrow$}
    & {SSIM$\uparrow$}
    & {LPIPS$\downarrow$}
    & {VFID$_\mathrm{I}\downarrow$}
    & {VFID$_\mathrm{R}\downarrow$} \\
    \hhline{-------}
    \XSolidBrush & 8.9187 & 0.2353 & 0.8898 & 0.0488 & 13.2985 & 0.2007 \\
    \Checkmark & \textbf{8.3623} & \textbf{0.1934} & \textbf{0.8922} & \textbf{0.0456} & \textbf{12.3640} & \textbf{0.1876} \\
    \bottomrule
  \end{tabular}
  \caption{Ablation of task-specific LoRA adaptation for the MLLM on the ViViD-S video benchmark.}
  \label{tab:ablation_lora_mllm}
\end{table*}
\begin{table*}[tb]
  \centering
  \setlength{\tabcolsep}{2mm}
  \begin{tabular}{l|cccc|cc}
    \toprule
    \multirow{2}{*}{{Method}} &
    \multicolumn{4}{c|}{{Paired}} &
    \multicolumn{2}{c}{{Unpaired}} \\
    \hhline{~|----|--}
    & {VFID$_\mathrm{I}\downarrow$}
    & {VFID$_\mathrm{R}\downarrow$}
    & {SSIM$\uparrow$}
    & {LPIPS$\downarrow$}
    & {VFID$_\mathrm{I}\downarrow$}
    & {VFID$_\mathrm{R}\downarrow$} \\
    \hhline{-------}
    No-Bridge-UniVVT & 9.2603 & 0.1995 & 0.8807 & 0.04920 & 14.0105 & 0.2256 \\
    \textbf{UniVVT} & \textbf{8.3623} & \textbf{0.1934} & \textbf{0.8922} & \textbf{0.0456} & \textbf{12.3640} & \textbf{0.1876} \\
    \bottomrule
  \end{tabular}
  \caption{Quantitative ablation results of the semantic bridge module on the ViViD-S video benchmark.}
  \label{tab:ablation_bridge_module}
\end{table*}
\paragraph{\textbf{Adapting the MLLM to video try-on.}}
The pretrained MLLM supplies strong multimodal priors, but its hidden representation is not explicitly optimized for the body--garment correspondence, edit localization, and temporal transfer required by VVT. We therefore adapt the MLLM with LoRA during Stage 1 semantic alignment and Stage 2 joint optimization. To isolate the contribution of this adaptation, we compare against a variant that freezes the MLLM while retaining the same trainable semantic bridge, generator, and optimization schedule. As shown in Table~\ref{tab:ablation_lora_mllm}, the frozen variant remains competitive, confirming that generic visual--language pretraining already provides a useful foundation. LoRA adaptation nevertheless improves every paired and unpaired metric: for example, paired VFID$_\mathrm{R}$ decreases from 0.2353 to 0.1934, while unpaired VFID$_\mathrm{I}$ decreases from 13.2985 to 12.3640. These consistent gains indicate that VVT requires more than generic garment recognition. Task-specific adaptation reorganizes the MLLM features around the complete editing context, enabling the task queries to jointly capture target appearance, source-person structure, spatial edit intent, and motion-aware correspondence. This produces a more actionable representation for the semantic bridge and reduces the burden on downstream modules to infer task structure from a fixed feature space. The progressive schedule further stabilizes this specialization: Stage 1 aligns the adapted representation to a frozen generator, and Stage 2 co-adapts perception and generation end to end. Thus, LoRA serves not merely as parameter-efficient fine-tuning, but as a mechanism for converting general multimodal knowledge into a task-specific semantic interface for video generation.
\begin{figure}[tb]
  \centering
  \includegraphics[width=\linewidth]{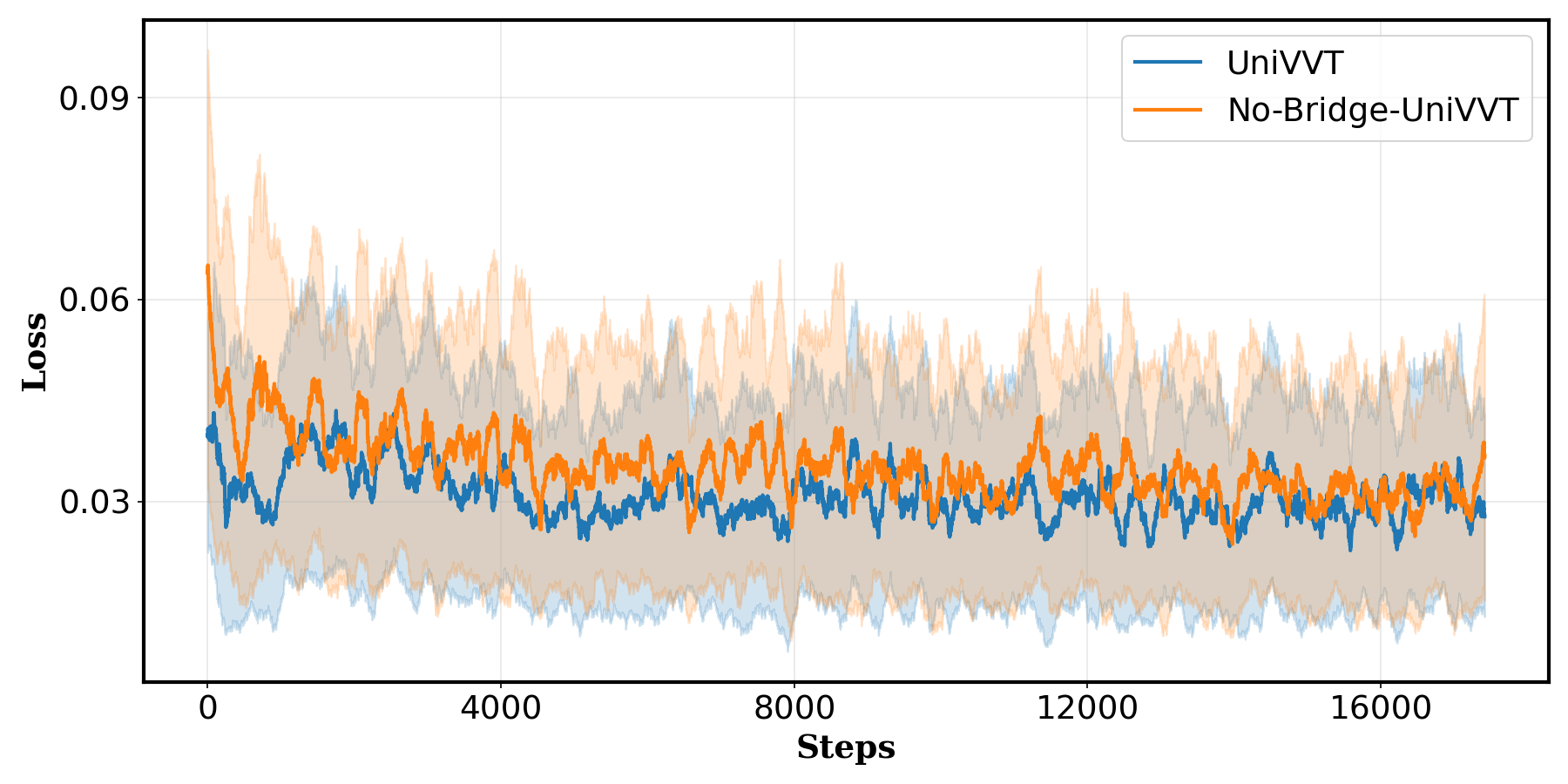}
    \caption{Stage 2 training-loss curves for UniVVT with and without the Semantic Bridge. The learned bridge accelerates convergence and consistently yields a lower optimization loss.}
    \label{fig:loss_compaire}
\end{figure}
\paragraph{\textbf{Learned alignment between perception and generation.}}
The MLLM and video DiT are pretrained independently and expose incompatible representation spaces. While $\mathbf{z}_{\text{task}}$ encodes multimodal editing intent, the DiT expects conditioning features whose statistics and semantics are compatible with its denoising dynamics. Resolving the dimensional mismatch alone is therefore insufficient. To test whether learned alignment is necessary, we construct \emph{No-Bridge-UniVVT}, which removes $\mathcal{P}_{\theta}$ and directly injects a zero-padded $\mathbf{z}_{\text{task}}$ while keeping all remaining components and the training schedule unchanged. As reported in Table~\ref{tab:ablation_bridge_module}, the Semantic Bridge improves every metric under both evaluation protocols. In particular, unpaired VFID$_\mathrm{I}$ decreases from 14.0105 to 12.3640 and VFID$_\mathrm{R}$ from 0.2256 to 0.1876, demonstrating stronger generalization beyond matched garment--person pairs. The paired gains in SSIM and LPIPS further show that this improvement is not limited to distribution-level realism, but also extends to structural preservation and perceptual fidelity. Figure~\ref{fig:loss_compaire} provides complementary optimization evidence: the full model converges faster and reaches a consistently lower Stage 2 training loss. Direct injection forces the generator to learn the VVT denoising objective while simultaneously interpreting features from an incompatible MLLM space. The bridge separates these demands by mapping $\mathbf{z}_{\text{task}}$ into generator-compatible tokens $\mathbf{c}_{\text{proj}}$, allowing the DiT to focus on spatially precise and temporally coherent synthesis. This interpretation also motivates the progressive training strategy: Stage 1 learns the cross-model interface against a frozen generator, and Stage 2 subsequently co-adapts perception, alignment, and generation. The Semantic Bridge is thus more than a dimensional projector; it is a lightweight learned interface that converts multimodal task understanding into effective generative control.

\begin{table*}[tb]
\vspace{-1em}
  \centering
  \setlength{\tabcolsep}{2mm}
  \renewcommand{\arraystretch}{1}
    \begin{tabular}{l|c|c|cccc}
      \toprule
      \multirow{2}{*}{Bridge Design} &\multirow{2}{*}{Params} & \multirow{2}{*}{FLOPs} & \multicolumn{4}{c}{Paired}   \\
      \hhline{~|~|~|----}
      &  &  & {VFID$_\mathrm{I}\downarrow$} & {VFID$_\mathrm{R}\downarrow$} & {SSIM$\uparrow$} & {LPIPS$\downarrow$}  \\
      \hhline{-------}
      12-layer Transformer & 1260.08M & 571.46G & 8.8922 & 0.2026 & 0.8869 & 0.0482  \\
      MLP &41.96M & 21.48G & 8.3623 & 0.1934  & 0.8922  & 0.0456   \\
      \bottomrule
\end{tabular}
  \caption{Ablation study on the bridge design.}
  \label{tab:ablation_bridge}
\end{table*}

\paragraph{\textbf{How much capacity does the semantic bridge need?}}
Having established the necessity of a learnable interface, we next investigate whether increasing its capacity further improves semantic alignment. Under the same training protocol and optimization budget, we compare two bridge topologies: a heavy 12-layer Transformer and the lightweight MLP adopted in UniVVT. As shown in Table~\ref{tab:ablation_bridge}, despite its substantially higher parameter and computational costs, the Transformer surprisingly performs slightly worse than the MLP. This result suggests that the Semantic Bridge need not relearn the complex multimodal reasoning already encoded in the task-aware representation $\mathbf{z}_{\text{task}}$; instead, its primary role is to perform a compact distributional transformation into the generator's conditioning space. Excessive capacity introduces redundant degrees of freedom and lengthens the optimization path across the perceiver--bridge--generator pipeline, potentially making coordinated adaptation more difficult. In contrast, the MLP offers sufficient expressiveness for semantic alignment while retaining a short and stable information pathway. These findings justify our lightweight design as both more effective and substantially more efficient, demonstrating that successful cross-model alignment relies on an appropriately constrained interface rather than maximal bridge capacity.
\section{Additional Quantitative Experiments}
\paragraph{Comparison with recent image-only methods.}
The main paper has already established the image try-on performance of UniVVT on standard benchmarks. Here, we broaden the comparison to recent image-only systems, including the general-purpose editors Qwen-Image-Edit~\cite{zhao2026qwen} and FLUX.2-klein~\cite{labs2025flux}, and the specialized try-on method UniFit~\cite{zhang2026unifit}. UniVVT is designed for video virtual try-on, but its mixed image--video formulation represents an image as a single-frame sequence, allowing the same architecture to operate in the image setting without modality-specific components. In contrast, all compared baselines are restricted to static images and do not support video try-on. Table~\ref{tab:more_comparison} therefore places a video-oriented unified model against recent methods developed specifically for image editing or image try-on, providing a stricter assessment of its cross-modal compatibility rather than repeating the image capability study from the main paper.

\begin{table*}[tb]
    \centering
    \setlength{\tabcolsep}{2mm}
      \begin{tabular}{l|c|cccc|cc}
        \toprule
        \multirow{2}{*}{{Method}} & 
        \multirow{2}{*}{\makecell{Supports\\video?}} & 
        \multicolumn{4}{c|}{{Paired}} & 
        \multicolumn{2}{c}{{Unpaired}} \\
        \hhline{~|~|----|--}
        & & {FID$\downarrow$} & {KID$\downarrow$} & {SSIM$\uparrow$} & {LPIPS$\downarrow$} & {FID$\downarrow$} & {KID$\downarrow$} \\
        \hhline{--------}
        Qwen-Image-Edit\cite{zhao2026qwen} & \textcolor{red}{\XSolidBrush} & 14.269 & 6.1090 & 0.6976 & 0.2688 & 13.5681 & 5.688  \\
        FLUX.2-klein\cite{labs2025flux} & \textcolor{red}{\XSolidBrush} & 11.503 & 4.5830 & 0.8304 & 0.1189 & 12.4746 & 4.1504  \\
        UniFit\cite{zhang2026unifit} & \textcolor{red}{\XSolidBrush} & 8.799 & 0.702 & \textbf{0.883} & 0.065  & / & /  \\
        \hhline{--------}
        \textbf{UniVVT} & \textcolor{blue}{\Checkmark}  & \textbf{5.348} & \textbf{0.3667}  & {0.8821}  & \textbf{0.0512}  & \textbf{9.014}  & \textbf{1.130}  \\
        \bottomrule
  \end{tabular}%
    
    \caption{Additional comparison with recent image-only editing and try-on methods on VITON-HD. UniVVT is designed for video virtual try-on and is the only model in this comparison that also supports the video setting.}
    \label{tab:more_comparison}
  \end{table*}

Table~\ref{tab:more_comparison} shows that the general-purpose editors perform substantially worse on virtual try-on, where the target garment must be reconstructed faithfully while person identity and non-garment content remain unchanged. The specialized UniFit model provides a considerably stronger comparison, yet UniVVT still achieves lower paired FID, KID, and LPIPS, with an essentially identical SSIM of 0.8821 versus 0.883. UniVVT also obtains the best reported FID and KID under the unpaired protocol. Crucially, these image-level results are achieved by a model developed for video try-on and equipped to model temporal dynamics; none of the competing methods supports the video task. UniVVT therefore does not obtain video capability at the expense of static-image quality. Instead, its unified formulation subsumes image try-on as a single-frame case, matching or surpassing dedicated image systems while retaining compatibility with video generation.
\section{Additional Qualitative Results}
\begin{figure}[tb]
  \centering
  \includegraphics[width=\linewidth]{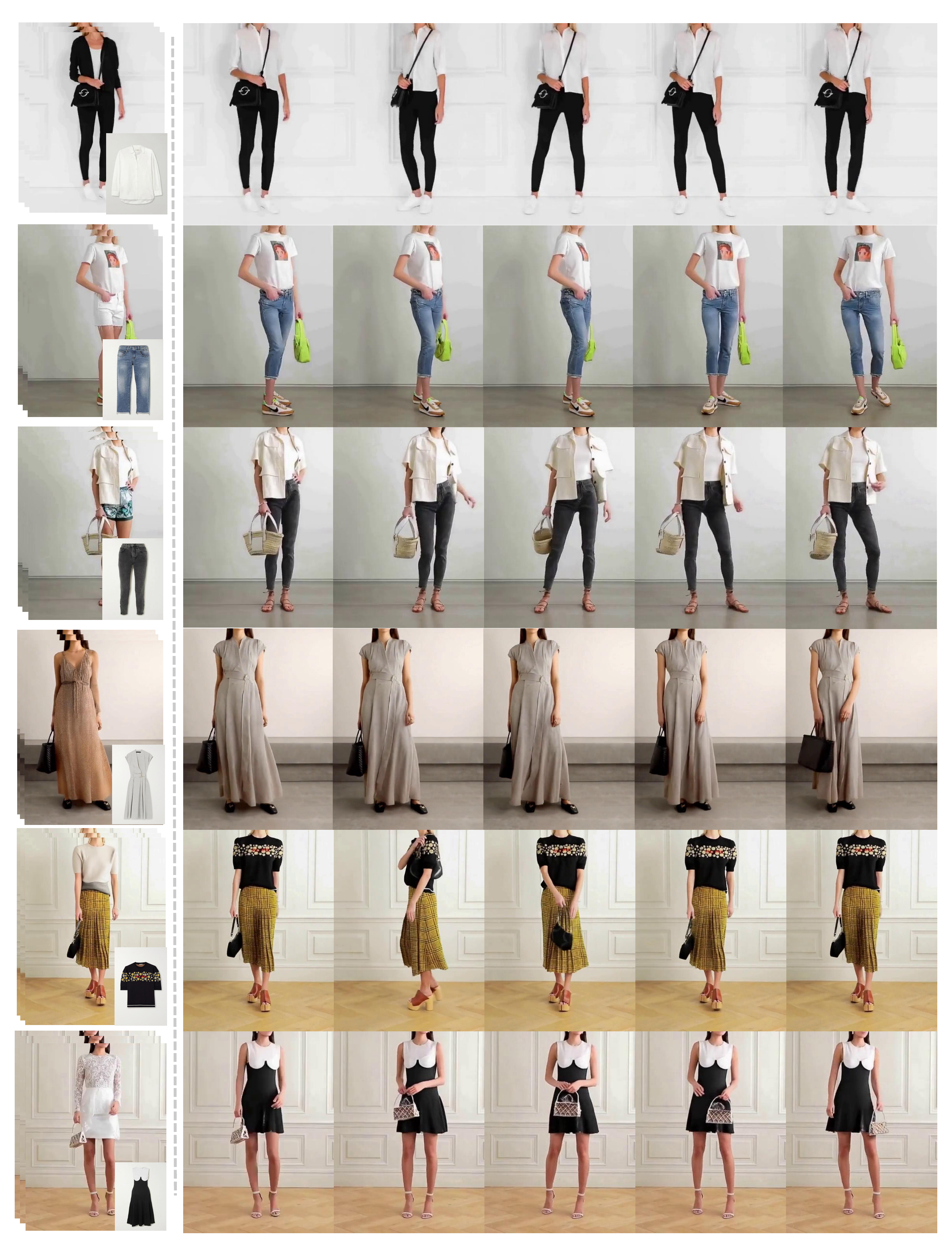}
    \caption{Additional video try-on results across garment categories, motion patterns, and occlusion conditions. The source sequence and target garment are shown on the left, followed by representative generated frames. UniVVT maintains garment appearance and structure over time while preserving the person and scene.}
    \label{fig:more_tryon_examples}
\end{figure}
\paragraph{Diverse video try-on scenarios.}
Figure~\ref{fig:more_tryon_examples} presents additional video results spanning upper-body garments, trousers, and dresses under varied poses, motions, and self-occlusions. UniVVT consistently transfers target color, texture, silhouette, and body coverage while preserving identity, pose, accessories, and background content. In particular, garment appearance remains stable when arms or handbags intermittently occlude the edited region and when the target silhouette differs markedly from the source clothing. These cases require the model to update garment--body correspondence as the person moves rather than applying a fixed spatial edit. Without masks, pose maps, or warped garments at inference, UniVVT infers this correspondence directly from the source video, target garment, and instruction. The resulting spatial precision and cross-frame consistency provide further evidence that task-aware semantic guidance can replace explicit geometric priors across diverse video try-on conditions.

\begin{figure}[tb]
  \centering
  \includegraphics[width=\linewidth]{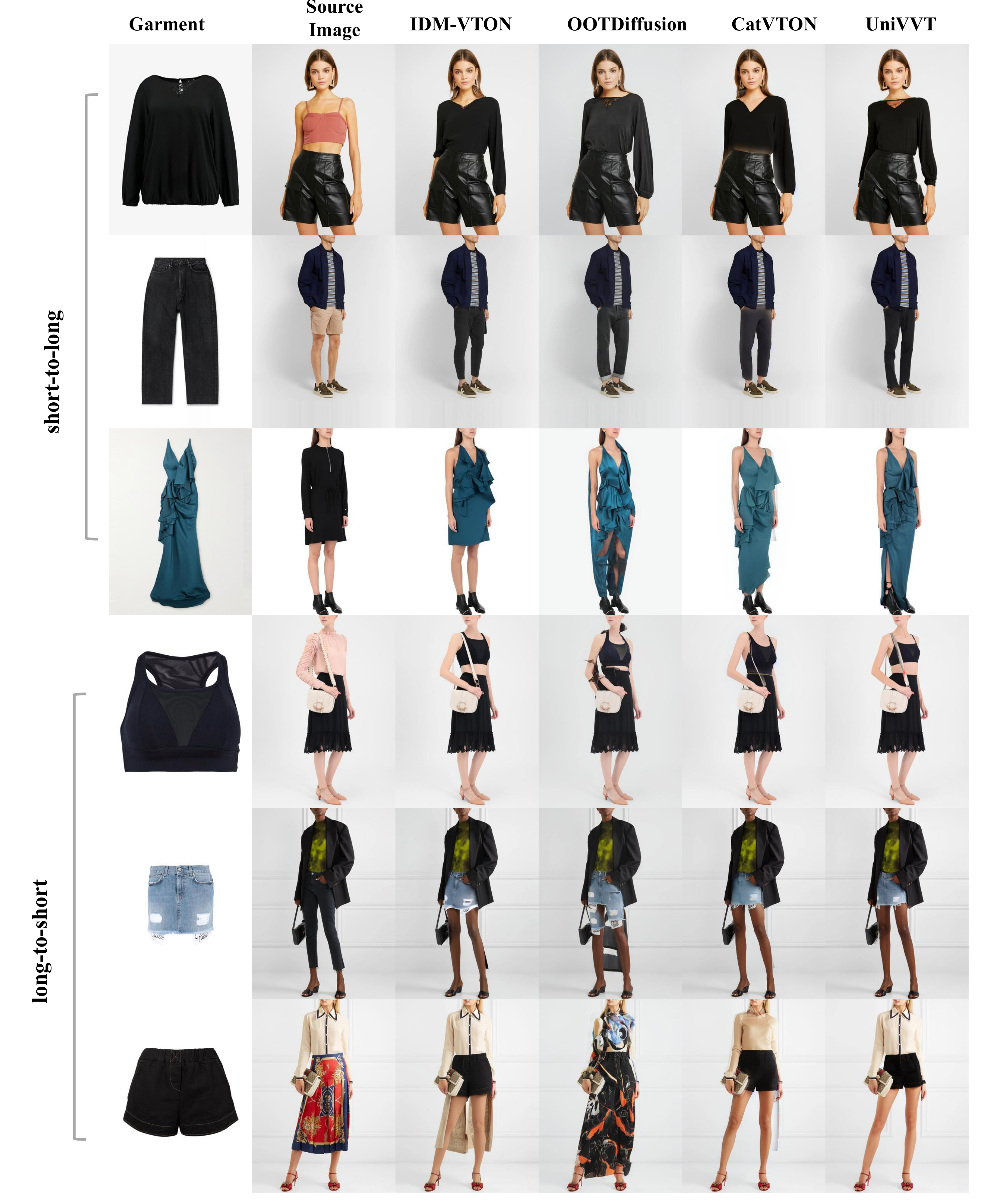}
    \caption{Image try-on under large changes in garment length and spatial coverage (top: short-to-long; bottom: long-to-short). Mask-conditioned baselines frequently inherit an incorrect editing extent from the source garment, whereas UniVVT adapts the edited region to the target and recovers more faithful silhouettes.}
    \label{fig:long_short}
\end{figure}

\paragraph{Bidirectional garment-length transfer.}
Figure~\ref{fig:long_short} examines image try-on under large changes in garment length and spatial coverage, with short-to-long transfers in the top three rows and long-to-short transfers in the bottom three. These cases directly stress the determination of the editing extent: a longer target must synthesize content beyond the source garment boundary, whereas a shorter target must expose and preserve regions previously covered by the source clothing. Mask-conditioned baselines often inherit their editable support from source-dependent preprocessing. Consequently, they truncate extended sleeves or hems, retain residual long-garment structures, or produce unnatural transitions at the target boundary. UniVVT instead infers the required spatial support jointly from the source person and target garment. It reconstructs long sleeves, trousers, and full-length dresses while also recovering compact tops, skirts, and shorts with cleaner boundaries and more faithful silhouettes. The bidirectional results demonstrate that implicit semantic guidance adapts the editing region to the target garment rather than constraining generation to a predefined source mask.
\section{Additional Visual Comparisons}
\begin{figure}[tb]
  \includegraphics[width=\linewidth]{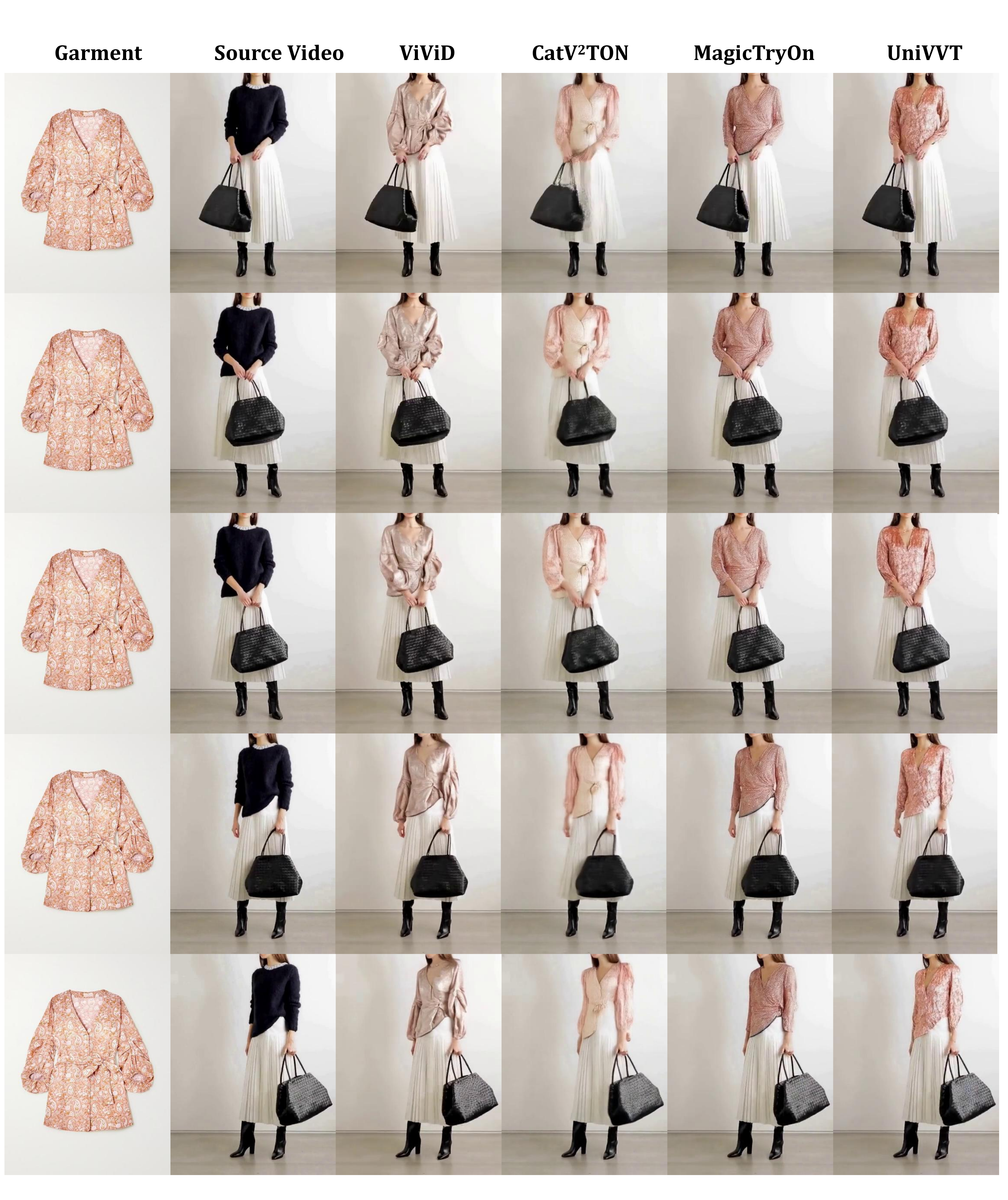}
  \caption{Fine-grained pattern preservation in video try-on.}
  \label{fig:compare_vivid_appendix_1}
\end{figure}

\begin{figure}[tb]
  \includegraphics[width=\linewidth]{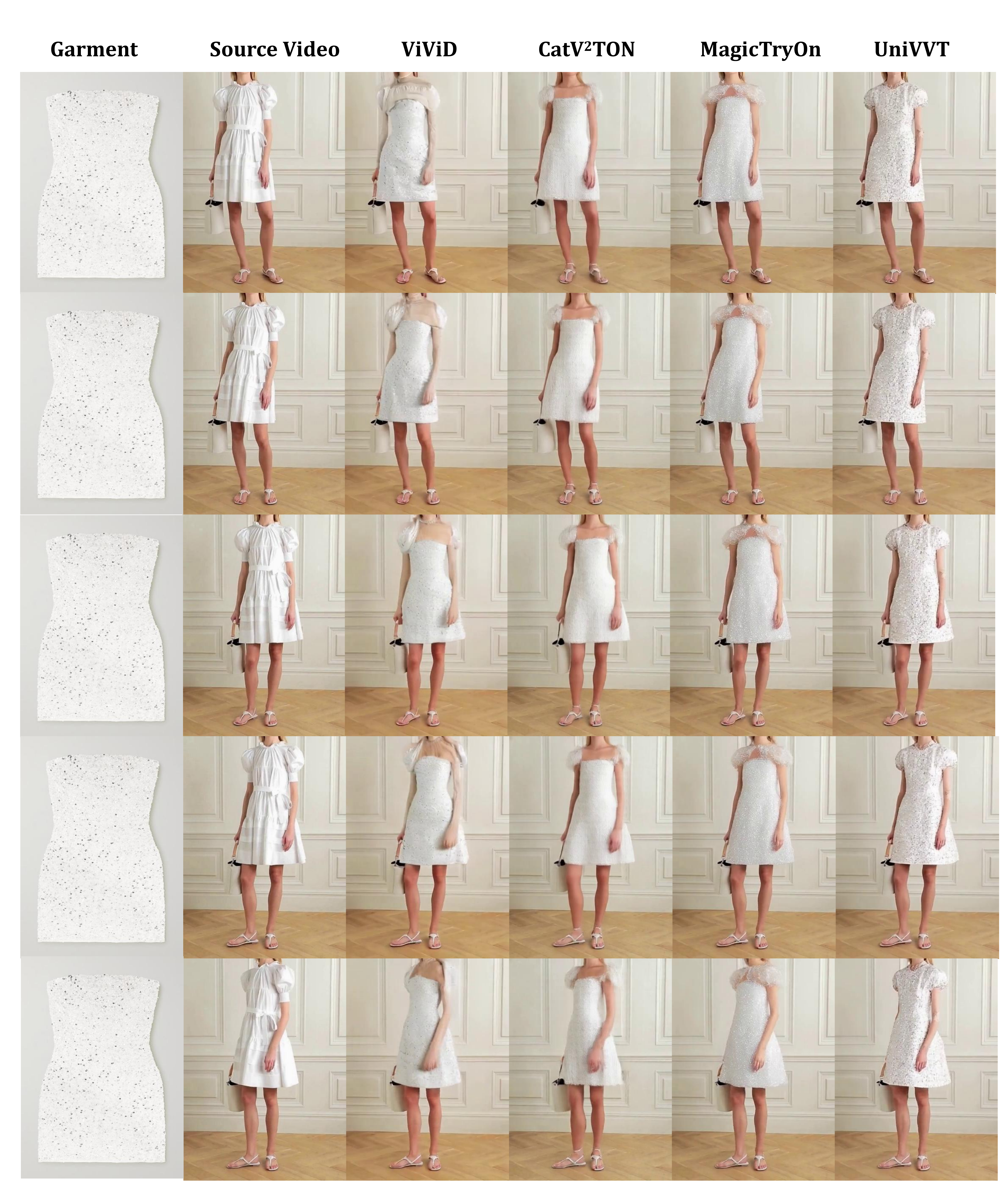}
  \caption{Subtle texture preservation in video try-on.}
  \label{fig:compare_vivid_appendix_3}
\end{figure}

\begin{figure}[tb]
  \includegraphics[width=\linewidth]{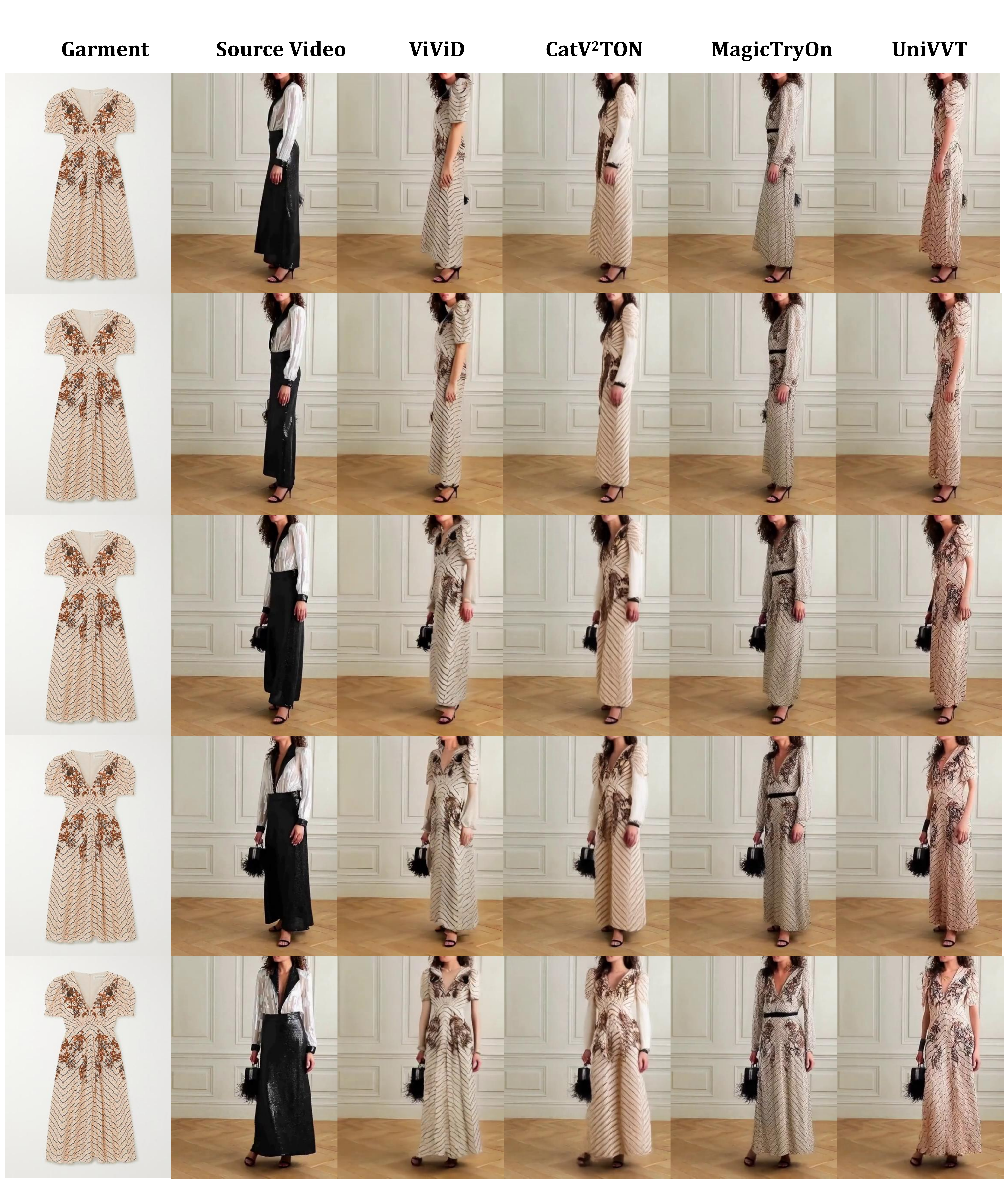}
  \caption{Pattern and color fidelity in video try-on.}
  \label{fig:compare_vivid_appendix_5}
\end{figure}

\begin{figure}[tb]
  \includegraphics[width=\linewidth]{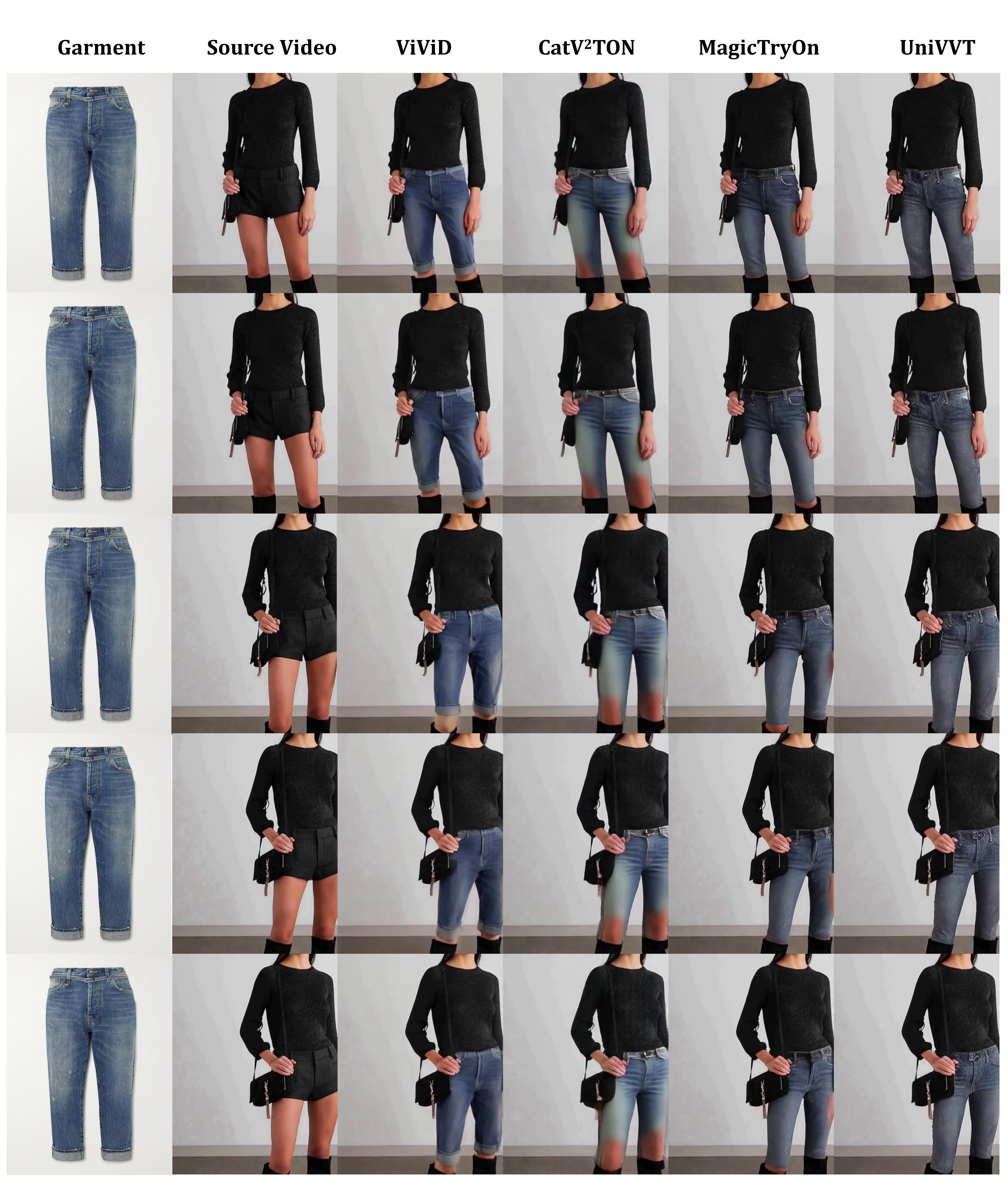}
  \caption{Complete trouser replacement in video try-on.}
  \label{fig:compare_vivid_appendix_2}
\end{figure}

\begin{figure}[tb]
  \includegraphics[width=\linewidth]{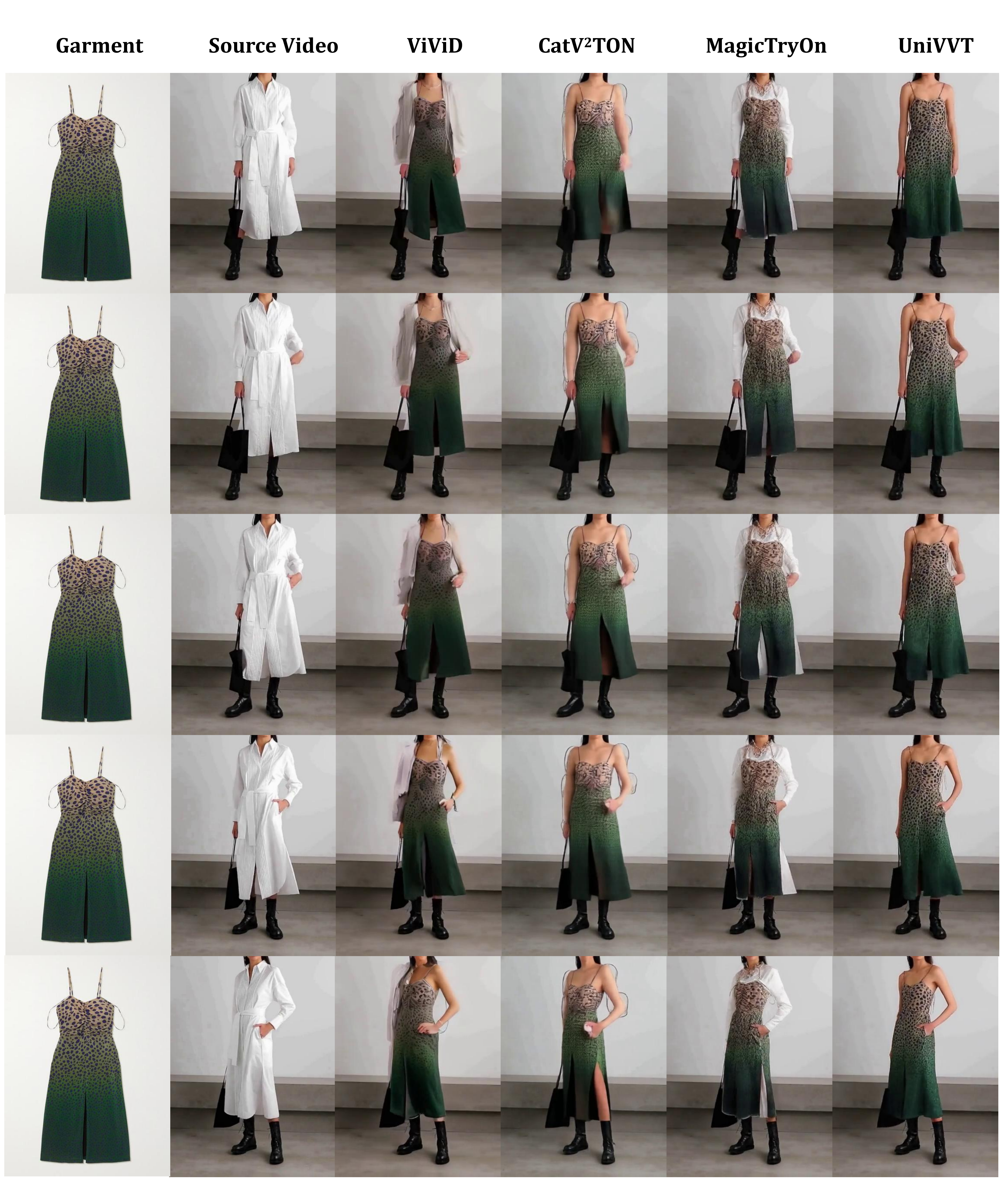}
  \caption{Source-garment leakage suppression in video try-on.}
  \label{fig:compare_vivid_appendix_4}
\end{figure}

\paragraph{Video virtual try-on.}
Figures~\ref{fig:compare_vivid_appendix_1}--\ref{fig:compare_vivid_appendix_5} compare UniVVT with open-source state-of-the-art video try-on methods. Figures~\ref{fig:compare_vivid_appendix_1}, \ref{fig:compare_vivid_appendix_3}, and \ref{fig:compare_vivid_appendix_5} highlight appearance fidelity: UniVVT reconstructs the distinctive paisley print, fine speckled texture, and structured brown patterns with sharper details and more accurate colors, whereas ViViD, CatV2TON, and MagicTryOn often smooth textures, shift colors, or omit characteristic motifs. Figures~\ref{fig:compare_vivid_appendix_2} and \ref{fig:compare_vivid_appendix_4} expose a complementary weakness of mask-conditioned baselines. Mask errors lead to truncated trousers, blurred leg boundaries, or residual white sleeves and panels from the source garment. By inferring the editing scope directly from the original multimodal inputs, UniVVT produces complete silhouettes, clean garment boundaries, and consistent appearance across frames.

\begin{figure}[tb]
  \includegraphics[width=\linewidth]{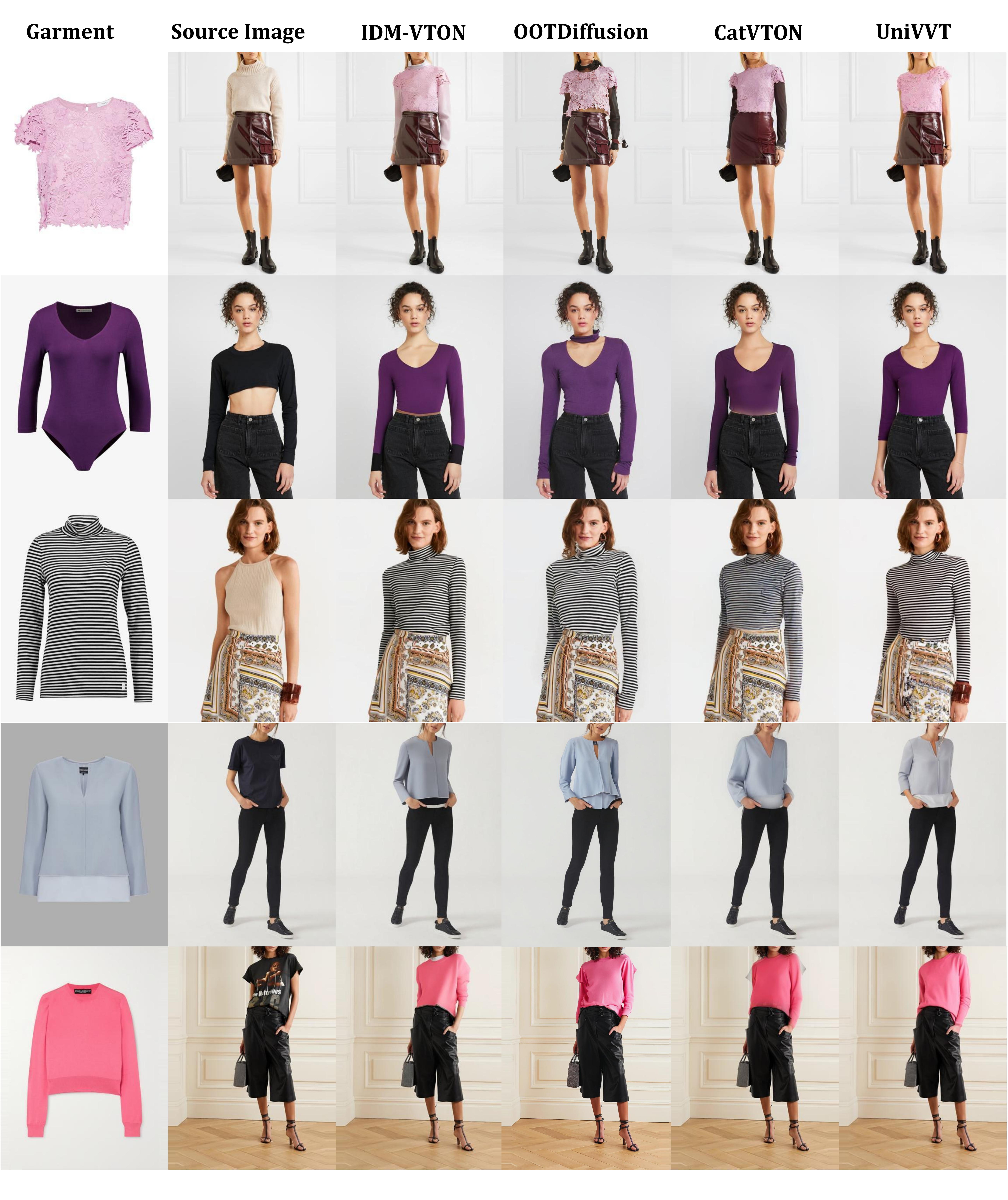}
  \caption{Upper-body image try-on comparison.}
  \label{fig:compare_vton_upper}
\end{figure}

\begin{figure}[tb]
  \includegraphics[width=\linewidth]{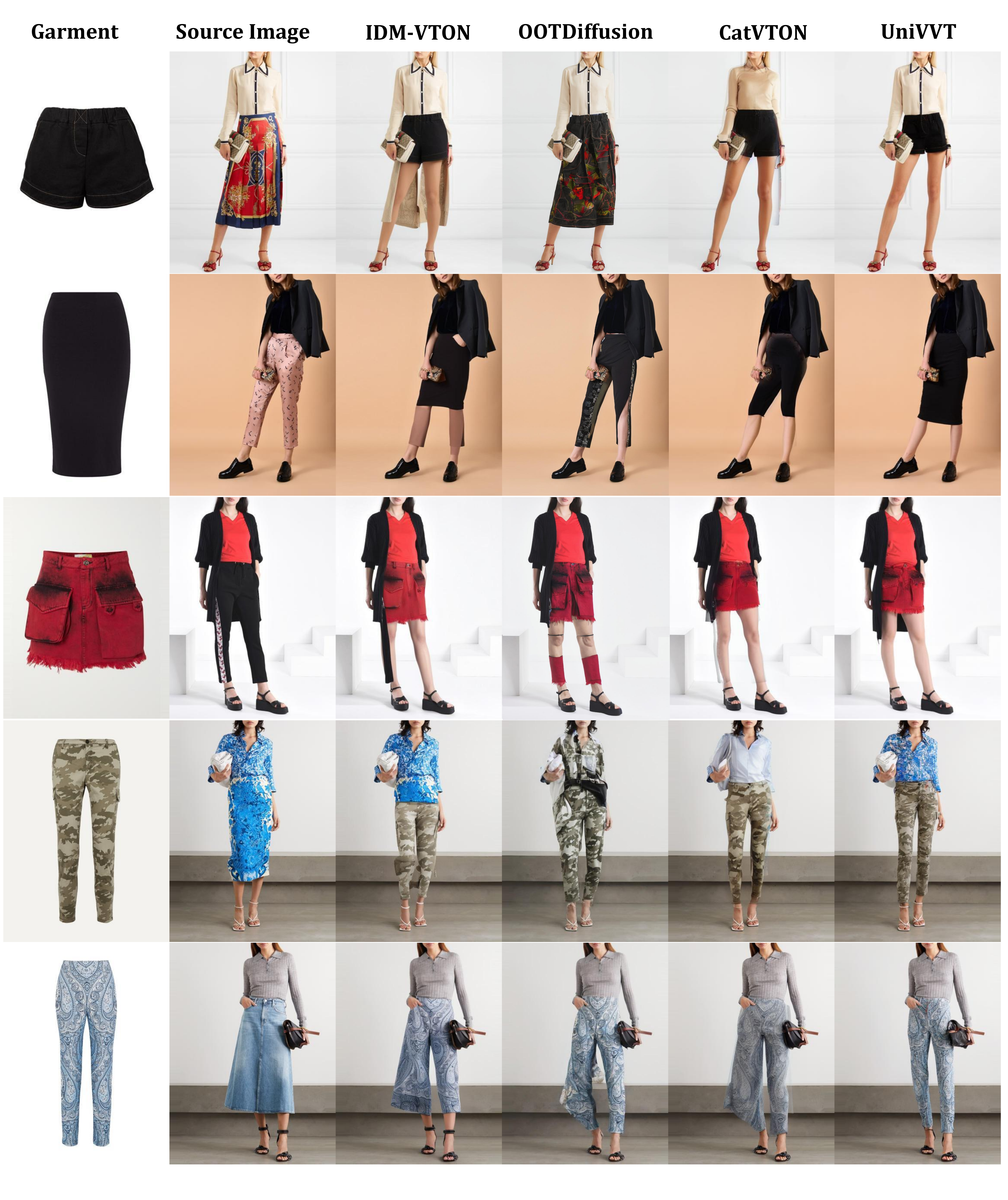}
  \caption{Lower-body image try-on comparison.}
  \label{fig:compare_vton_lower}
\end{figure}

\begin{figure}[tb]
  \includegraphics[width=\linewidth]{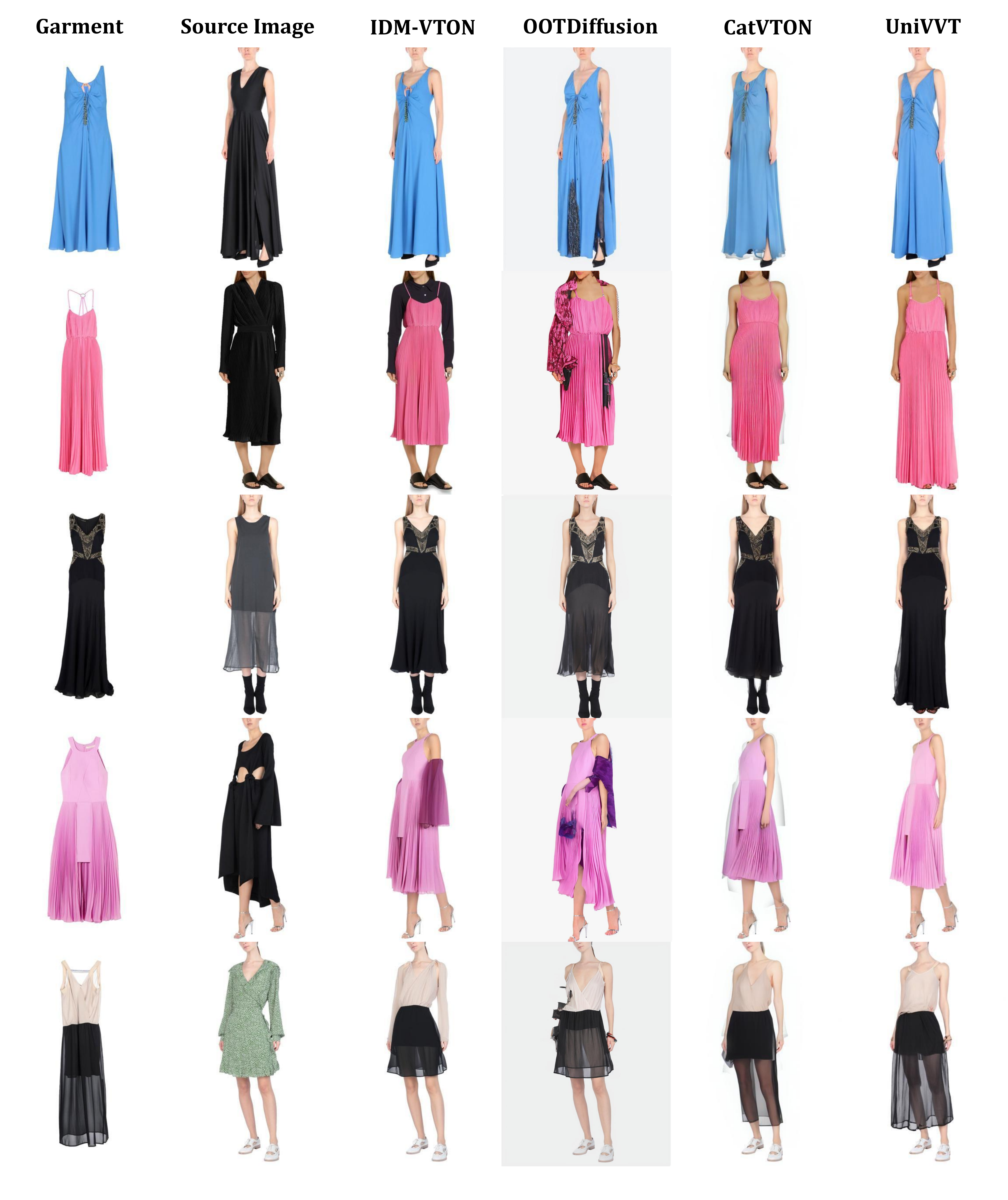}
  \caption{Dress image try-on comparison.}
  \label{fig:compare_vton_dresses}
\end{figure}

\paragraph{Image virtual try-on.}
In the image virtual try-on setting, Figures~\ref{fig:compare_vton_upper}, \ref{fig:compare_vton_lower}, and \ref{fig:compare_vton_dresses} compare UniVVT on upper-body garments, lower-body garments, and dresses. UniVVT more faithfully preserves neckline and sleeve configurations, garment length and drape, and localized details such as lace, stripes, cargo pockets, and camouflage patterns. Meanwhile, person identity, pose, accessories, and background remain intact. These comparisons demonstrate that UniVVT transfers its unified semantic modeling capability effectively to image virtual try-on, achieving precise garment reconstruction and localized editing across diverse garment categories without an image-specific architecture.

\section{Limitations and Future Directions}
\label{sec:limitation}
Despite its strong performance, UniVVT retains three limitations that motivate further study.

\paragraph{Temporal stability under extreme motion.}
Fast body motion, abrupt viewpoint changes, and severe occlusion can induce local texture drift, discontinuous garment patterns, or short-term flicker. The task-aware representation captures global editing intent, but does not explicitly enforce dense correspondence between every pair of adjacent frames. Stronger temporal memory, motion-aware feature propagation, and consistency objectives may therefore improve fine-grained stability in highly dynamic sequences.

\begin{figure}[t]         
  \includegraphics[width=\linewidth]{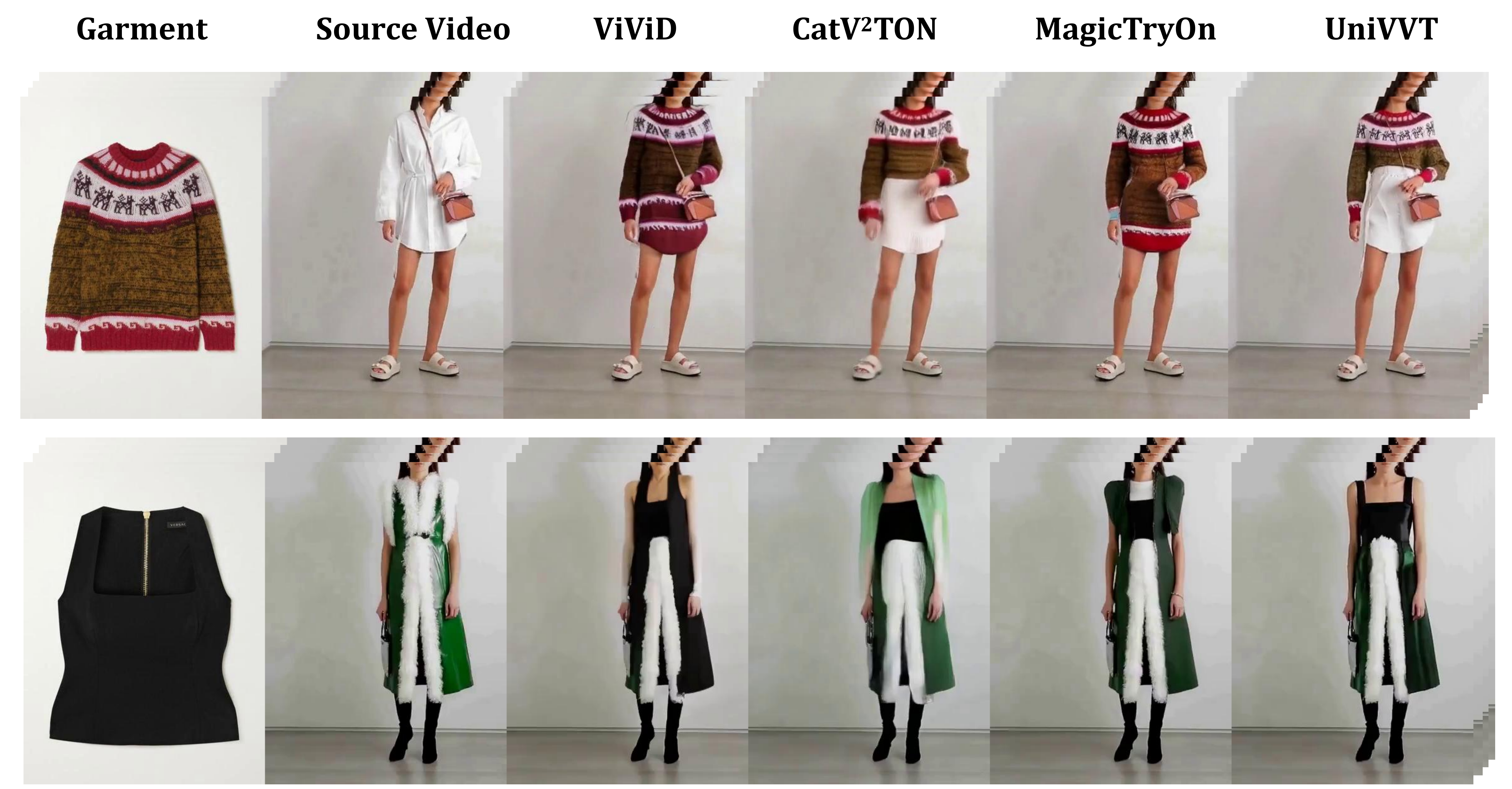}
    \caption{Cross-category failure cases with ambiguous editing extent. Replacing a dress with an upper-body garment requires determining which previously covered regions should be preserved, removed, or synthesized; none of the evaluated methods resolves this ambiguity reliably.}
  \label{fig:fail_case_2}
\end{figure}

\paragraph{Ambiguity in cross-category transfer.}
Our training data primarily support garment replacement within the same semantic category. Cross-category editing, such as replacing a dress with a top, changes not only garment appearance but also the intended body coverage. As illustrated in Figure~\ref{fig:fail_case_2}, the target garment alone does not specify whether regions previously covered by the dress should be preserved, removed, or newly synthesized; none of the evaluated methods resolves this ambiguity reliably. Extending training data to cross-category transformations and introducing finer-grained instructions about the desired outfit composition and editing extent are promising directions.

\paragraph{Inference efficiency.}
UniVVT eliminates costly frame-wise geometric preprocessing, but iterative diffusion denoising remains the dominant runtime bottleneck. Consequently, the current model is not yet suitable for real-time interactive applications. Few-step distillation, accelerated sampling, and more efficient video-generation backbones could reduce latency while preserving garment fidelity and temporal coherence.



\end{document}